\documentclass[10pt,journal,compsoc]{IEEEtran}
\ifCLASSOPTIONcompsoc
  \usepackage[nocompress]{cite}
\else
  \usepackage{cite}
\fi
\ifCLASSINFOpdf
\else
\fi
\usepackage{times}
\usepackage{epsfig}
\usepackage{graphicx}
\usepackage{amsmath}
\usepackage{amssymb}

\usepackage{hyperref}       
\usepackage{url}            
\usepackage{booktabs}       
\usepackage{amsfonts}       
\usepackage{nicefrac}       
\usepackage{microtype}      
\usepackage{floatrow}
\newfloatcommand{capbtabbox}{table}[][\FBwidth]
\usepackage{xcolor}

\usepackage{subfigure}
\usepackage{enumitem}
\usepackage{algorithm}
\usepackage{algorithmic}
\usepackage{comment}
\usepackage{subfigure}
\usepackage{booktabs}
\usepackage{array}
\usepackage{pifont}
\usepackage{graphbox}
\usepackage{epstopdf}
\usepackage{sidecap}
\usepackage{wrapfig}

\usepackage{bm}
\usepackage{multirow}
\usepackage{wrapfig}
\usepackage{xcolor}
\usepackage{color, colortbl}
\definecolor{Gray}{gray}{0.9}
\definecolor{White}{rgb}{1,1,1}
\usepackage[english]{babel}

\begin{document}
%
\title{Conditional Temporal Variational AutoEncoder for Action Video Prediction}
%
%
%
%

\author{Xiaogang Xu, Yi Wang, Liwei Wang, Bei Yu, and Jiaya Jia,~\IEEEmembership{Fellow,~IEEE}
	\IEEEcompsocitemizethanks{\IEEEcompsocthanksitem Xiaogang Xu, Yi Wang, Liwei Wang, Bei Yu and Jiaya Jia are with the Department of Computer Science and Engineering, The Chinese University of Hong Kong (CUHK).\protect\\
		E-mail: \{xgxu, yiwang, lwwang, byu, leojia\}@cse.cuhk.edu.hk
}}

\IEEEtitleabstractindextext{%
\begin{abstract}
To synthesize a realistic action sequence based on a single human image, it is crucial to model both motion patterns and diversity in the action video. This paper proposes an Action Conditional Temporal Variational AutoEncoder (ACT-VAE) to improve motion prediction accuracy and capture movement diversity. ACT-VAE predicts pose sequences for an action clips from a single input image. It is implemented as a deep generative model that maintains temporal coherence according to the action category with a novel temporal modeling on latent space. Further, ACT-VAE is a general action sequence prediction framework. When connected with a plug-and-play Pose-to-Image (P2I) network, ACT-VAE can synthesize image sequences. Extensive experiments bear out our approach can predict accurate pose and synthesize realistic image sequences, surpassing state-of-the-art approaches. Compared to existing methods, ACT-VAE improves model accuracy and preserves diversity.
\end{abstract}

\begin{IEEEkeywords}
Temporal Variational AutoEncoder, Action Modeling, Temporal Coherence, Adversarial Learning.
\end{IEEEkeywords}}

\maketitle

\IEEEdisplaynontitleabstractindextext

%
\IEEEpeerreviewmaketitle

\IEEEraisesectionheading{\section{Introduction}\label{sec:introduction}}

\begin{figure}[t]
	\centering
	\newcommand\widthface{0.3}
	\resizebox{0.9\linewidth}{!}{
		\begin{tabular}{c@{\hspace{1.0mm}}c@{\hspace{1.0mm}}c}
			\includegraphics[align=c, width=\widthface\textwidth]{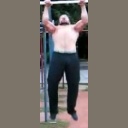} &
			\includegraphics[align=c, width=\widthface\textwidth]{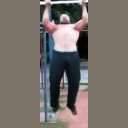} &
			\includegraphics[align=c, width=\widthface\textwidth]{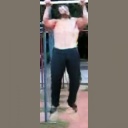} \\
			& \multicolumn{2}{c}{$\underbrace{\quad \quad \quad \quad \quad \quad \quad \quad}$}\\
			{Input}&\multicolumn{2}{c}{Input label I: pull up}\\
			&&\\
			&
			\includegraphics[align=c, width=\widthface\textwidth]{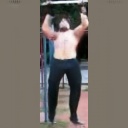} &
			\includegraphics[align=c, width=\widthface\textwidth]{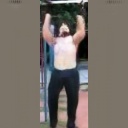}\\
			& \multicolumn{2}{c}{$\underbrace{\quad \quad \quad \quad \quad \quad \quad \quad}$}\\
			&\multicolumn{2}{c}{Input label II: squat}\\
	\end{tabular}}
	\caption{Given an input image, our method can synthesize future sequences and control their action types with the input action label.}
	\label{fig:teaser}
\end{figure}

\IEEEPARstart{H}{uman} action video prediction aims to generate future human action from single or multiple input human images \cite{yunji_neurips_2019,villegas2017learning,wichers2018hierarchical,zhao2018learning}. 
This topic is actively studied recently, for its importance to understand and improve human motion modeling and benefit in a variety of video applications, e.g, motion re-target \cite{aberman2019learning,villegas2018neural}.
In this work, we focus on synthesizing image sequences from a single image and controlling their action types via the input of action labels \cite{yunji_neurips_2019}, as shown in Fig. \ref{fig:teaser}. 

Due to the diversity in human motion, action video prediction is highly ill-posed with multiple possible solutions. Conventional deterministic models utilizing regression are useful, but over-smooth image sequences may be produced \cite{finn2016unsupervised,jia2016dynamic,kalchbrenner2017video,villegas2017decomposing}, giving mean estimation of future action. Recent deep generative approaches alleviate this problem by using Generative Adversarial Networks (GAN) \cite{cai2018deep,wichers2018hierarchical}, Variational AutoEncoder (VAE) \cite{yunji_neurips_2019,Prediction-ECCV-2018}, and Variational Recurrent Neural Network (VRNN) \cite{castrejon2019improved,denton2018stochastic} to model motion diversity explicitly.
Methods of \cite{cai2018deep,wichers2018hierarchical,yunji_neurips_2019,Prediction-ECCV-2018} use latent variables with identical independent distributions to capture motion patterns and diversity in every frame. Without temporal coherence among latent variables, action video prediction accuracy is bounded. Meanwhile, works of \cite{castrejon2019improved,denton2018stochastic} introduced unitive temporal coherence for all actions while ignored the distinction among different action categories.

In this paper, we treat the human pose as the high-level structure for action video prediction, and the predicted pose sequences are utilized as guidance for the synthesis of image sequences. This setting can avoid the interference of action-irrelevant appearance \cite{villegas2017learning,zhao2018learning} and thus usually outperform strategies of directly hallucinating images.

\begin{figure*}[t]
	\begin{center} 
		\includegraphics[width=1.0\linewidth]{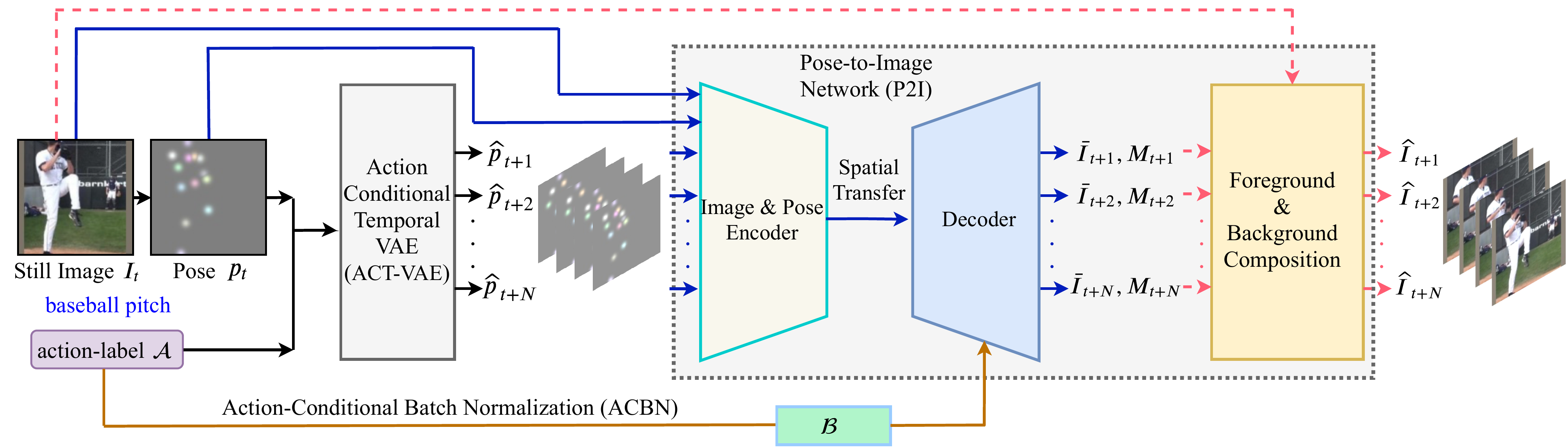}
	\end{center}
	\caption{Our framework for human action video prediction from a still image. It consists of two modules: ACT-VAE and P2I networks. 
		ACT-VAE can generate future pose sequences with novel temporal modeling on latent variables to achieve the individual temporal coherence (its structure is shown in Fig.~\ref{fig:vae}).
        ACT-VAE can further synthesize image sequences by connecting it with the plug-and-play P2I network.}
	\label{fig:framework}
\end{figure*}

To achieve modeling for the human pose, we propose Action Conditional Temporal Variational AutoEncoder (ACT-VAE) to describe the motion patterns and diversity, individually maintaining temporal coherence for each action category (so-called ``individual temporal coherence").
It is built upon a distinctive Recurrent Neural Network (RNN) \cite{mikolov2010recurrent,greff2016lstm} to maintain such coherence.
Similar to \cite{yunji_neurips_2019,yang2018pose}, we employ human key points as the representation of pose, and ACT-VAE predicts key points of the future pose sequence, based on the pose of the input image as well as an action label, as shown in Fig.~\ref{fig:framework}. 
We introduce action labels to the input and intermediate states of RNN for explicitly controlling what action to generate. 
Besides of the individual temporal coherence, compared with existing approaches \cite{yunji_neurips_2019,castrejon2019improved,wichers2018hierarchical,yuan2020dlow}, we incorporate novel temporal modeling on latent variables into ACT-VAE that improves motion prediction accuracy.
It updates the latent variable at each time step via the previous action features and the latent variables.
Moreover, extensive experiments validate its notable precision improvement for forecasting and comparable diversity in prediction with state-of-the-art methods \cite{yunji_neurips_2019,castrejon2019improved,yuan2020dlow}.

Furthermore, ACT-VAE can synthesize image sequences by connecting it with a plug-and-play network that maps pose to images.
To this, we design a Pose-to-Image (P2I) network to convert the predicted pose sequence from ACT-VAE into the image sequence with realistic appearances. 
To improve the synthesis, we explicitly disentangle the foreground part from the image sequence via an attention mechanism, and enhance synthesized results further by introducing action conditional batch normalization (ACBN) to the P2I network.
Extensive experiments on Penn-action dataset \cite{zhang2013actemes} and Human3.6M dataset \cite{ionescu2013human3} show the effectiveness of our method.
Our overall contribution is threefold.
\begin{itemize}
	\item We explicitly model individual temporal coherence for human action video prediction of diverse action types.
	\item We build ACT-VAE with novel temporal modeling on latent variables, improving the accuracy of action video prediction to a new level and simultaneously keep comparable diversity with existing methods.
	\item We show that ACT-VAE is very general and is applicable to synthesize plausible videos by connecting it with a plug-and-play P2I network. Significantly, our framework is flexible to generate various action types from single input, through controlling action labels.
\end{itemize}

\section{Related Work}
Some existing works for human action video prediction adopt deterministic models that directly minimize the distance
between the synthesized action frames and the real frames, to produce deterministic image sequences \cite{finn2016unsupervised,jia2016dynamic,yoo2017variational,kalchbrenner2017video,villegas2017learning,zhao2018learning,wang2019memory,kwon2019predicting,guen2020disentangling} or future pose \cite{li2018convolutional,zhao2018learning,wang2019imitation,guo2019human,gopalakrishnan2019neural,mao2019learning,cai2020learning,mao2020history,piergiovanni2020adversarial}.
Their corresponding performances are generally limited since their results may converge to the average of possible outcomes.
To achieve more realistic and dynamic predictions, recent methods employ deep generative models, including VAE \cite{kingma2013auto}, VRNN \cite{chung2015recurrent} and GAN \cite{goodfellow2014generative}.

GAN-based approaches extend the structure of vanilla GAN into the sequential one \cite{wichers2018hierarchical,lee2018stochastic,cai2018deep,mathieu2015deep}. 
Cai et. al. \cite{cai2018deep} predicted human pose sequences with a latent variable in adversarial learning.
The basic idea is to construct a discriminator to classify the realness of synthesized sequences and corresponding real ones. It updates the generator to pass the discriminator with good-quality generated sequences.
On the other hand, using VAE \cite{yunji_neurips_2019, Prediction-ECCV-2018,lee2018stochastic,yan2018mt,babaeizadeh2017stochastic,kumar2019videoflow,razavi2019preventing,aliakbarian2020stochastic} can also achieve promising performance.
Kim et al.~\cite{yunji_neurips_2019} extended VAE with RNN structure, and set a common latent variable for predicting overall time steps.
Lee et al.~\cite{lee2018stochastic} utilized multiple latent variables with identical distribution for prediction at each time step during inference.
Babaeizadeh et al.~\cite{babaeizadeh2017stochastic} modeled motion patterns and diversity with a single set of fixed latent variables for prediction. These strategies do not consider temporal coherence during inference.

Video prediction can also use VRNN \cite{castrejon2019improved,denton2018stochastic,minderer2019unsupervised}.
Denton et al.~\cite{denton2018stochastic} proposed a VRNN framework with a learned prior for inference.
Castrejon et al.~\cite{castrejon2019improved} improved the performance by extending hierarchical structures for latent variables of VRNN. These approaches do not maintain individual temporal coherence for each action category, and thus are different from our work. Besides, our framework's modeling on latent variables varies from the current VRNN.

One crucial issue about human action video prediction is the use of structural information, i.e., human pose. Some existing works \cite{srivastava2015unsupervised,xu2018structure,oliu2018folded,tulyakov2018mocogan,wichers2018hierarchical,lee2018stochastic,cai2018deep} directly synthesized action frames from networks and achieved success on simple datasets with low motion variance and image resolution. With the advance in pose-guided image generation \cite{zhu2019progressive,neverova2018dense,ma2017pose,Siarohin_2019_CVPR,Siarohin_2019_NeurIPS}, recent methods favored a two-stage strategy to generate pose sequences firstly and then use them as conditions to hallucinate image sequences \cite{Prediction-ECCV-2018,yunji_neurips_2019,zhao2018learning,villegas2017learning,wang2018every,yang2018pose,walker2017pose}.

\begin{figure}[!t]
	\centering 
	\includegraphics[width=1.0\linewidth]{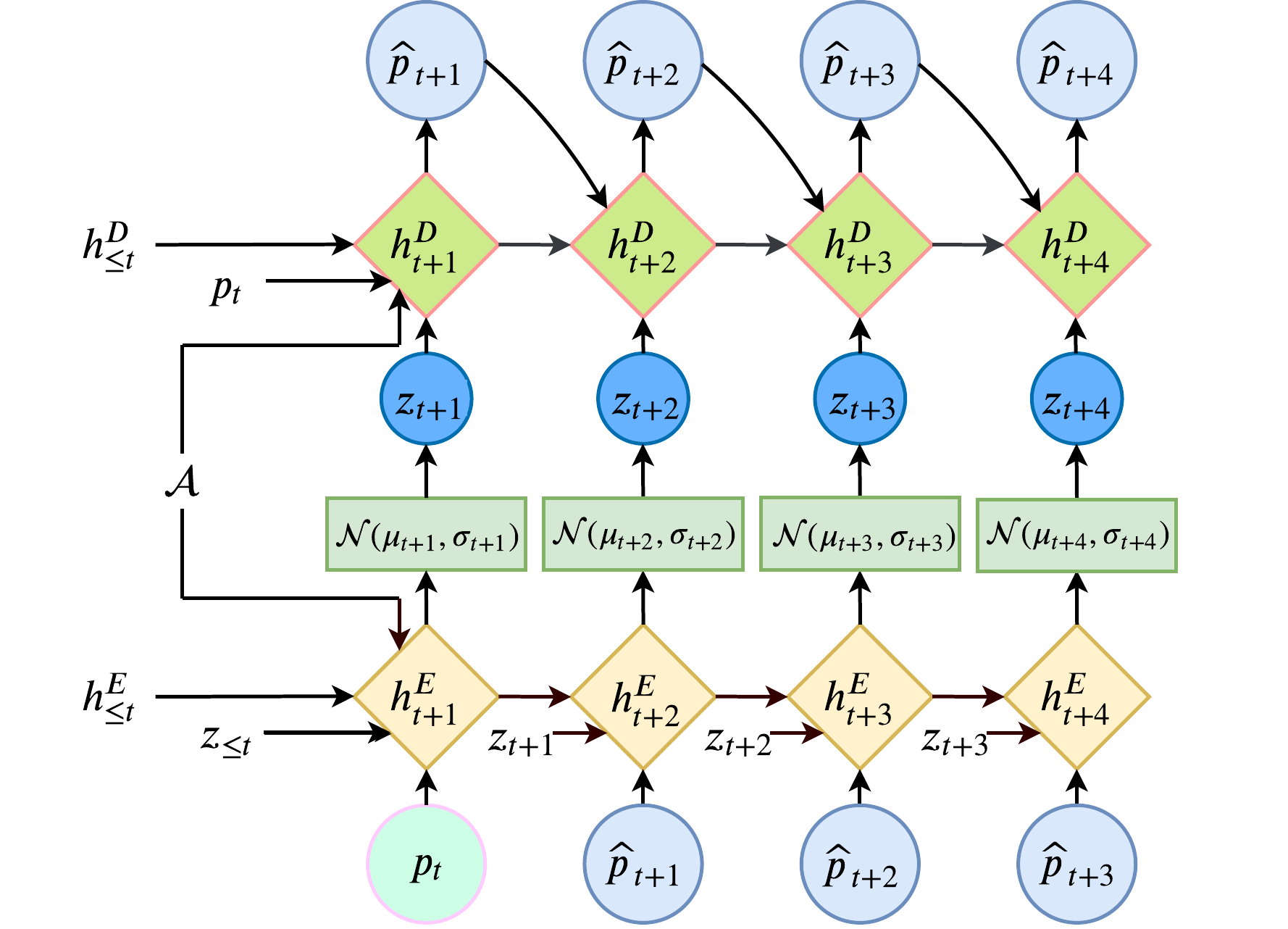}
	\caption{The architecture of ACT-VAE to synthesize future pose sequence based on an initial human pose $p_t$ and an action label $\mathcal{A}$. ACT-VAE consists of the encoder $E$ and the decoder $D$, which are both implemented with LSTM networks.
		Four future frames from the time index $t$ are generated here, and ACT-VAE can indeed synthesize the pose sequence with arbitrary length. Moreover, $h^E_{\leq t}$ and $h^D_{\leq t}$ are set as the initial hidden state of $E$ and $D$. In addition, we feed action labels into ACT-VAE by concatenating them with other input variables of LSTM at each time step.}
	\label{fig:vae}
\end{figure}

\section{Method}
Following the task setting of \cite{yunji_neurips_2019}, our model predicts human actions by synthesizing $N$ future RGB frames $\{\widehat{I}_{t+i}\}_{i=1,...,N}$ for an initial image $I_t \in \mathbb{R}^{H \times W \times 3}$ with its target action label $\mathcal{A} \in \mathbb{R}^{C}$ (in one-hot vector form) as
\begin{equation}
\{\widehat{I}_{t+i}\}_{i=1,...,N} = \mathcal{F}(I_t | \mathcal{A}),
\end{equation}
where $\mathcal{F}$ is the desired action video prediction model, $I$ denotes a real image with pose $p$, and $\widehat{I}$ denotes a synthesized image with pose $\widehat{p}$. $t$ and $t+i$ index time in a video. $H$ and $W$ denote image height and width respectively. $N$ and $C$ are the length of synthesized frames and the number of action categories to be modeled. 

Sequential action modeling should be independent of object appearance and background. To this end, we propose a framework consisting of two modules, which are ACT-VAE and P2I networks, as shown in Fig.~\ref{fig:framework}.
With the pose of the initial input image $I_t$ and the target action label $\mathcal{A}$, ACT-VAE generates pose sequences in key point form.
ACT-VAE further produces realistic videos by connecting it with the P2I network that is a plug-and-play module.

\subsection{Action Conditional Temporal VAE}
Given an initial image $I_t$, its initial pose $p_t$ (we employ the setting of \cite{villegas2017learning,zhao2018learning} to set $p_t$ for $I_t$), and target action label $\mathcal{A}$, our proposed ACT-VAE predicts future pose sequence as
\begin{equation}
\{\widehat{p}_{t+i}\}_{i=1,...,N} = \mathcal{T}(p_t | \mathcal{A}),
\end{equation}
where $\mathcal{T}$ is ACT-VAE. Both $p_t$ and $\widehat{p}_t$ can be represented in the form of key point coordinate values \cite{yunji_neurips_2019}.

The design of ACT-VAE is motivated by the observation that a particular kind of action should have a distinctive primary motion pattern. Meanwhile, it may exhibit diverse local details for different persons as the diversity of motion. 
For example, batting is a standard motion pattern in baseball while everybody's batting differs from each other a bit.
Such motion pattern and regional diversity are supposed to be temporally correlated for realism, and each action category should have its individual temporal coherence.

Therefore, unlike previous VAE-based video prediction methods~\cite{yunji_neurips_2019,Prediction-ECCV-2018} that take identical latent variables as the condition for the generation across all time steps, we equip VAE with the property of temporal coherence for various action categories.
As shown in Fig.~\ref{fig:vae}, ACT-VAE is different from conventional VAE by modeling the temporally correlated latent variable $\{ z_{t+i}\}_{i=1,...,N}$ and pose $\{ \widehat{p}_{t+i}\}_{i=1,...,N}$ with a distinctive recurrent structure, and using different action categories as the condition. 
Especially, ACT-VAE models the latent variable at each time step with features of both previous pose and latent variables.  
For variable $y$, $y_{<t}$ denotes the sequence $\{y_{t'}\}_{t'<t}$, and $y_{\leq t}$ represents $\{y_{t'}\}_{t'\leq t}$. 

\subsubsection{Structure of ACT-VAE}
As shown in Fig.~\ref{fig:vae}, ACT-VAE has an encoder $E$ and a decoder $D$ both in recurrent manner using LSTM \cite{greff2016lstm}.
The encoder $E$ is to sample latent variable $\{z_{t+i}\}_{i=1,...,N}$, if the initial input pose of ACT-VAE is denoted as $p_t$. In such process, the sampling of $z_{t+i}$ is implemented by modeling joint posterior distribution of $z_{<t+i}$ and $\widehat{p}_{<t+i}$ conditioned by the action label $\mathcal{A}$ as
\begin{equation}
\begin{split}
(\mu_{t+i}, \sigma_{t+i}, h^{E}_{t+i}) &= E( h^{E}_{t+i-1}, z_{t+i-1},\widehat{p}_{t+i-1} |\mathcal{A}), \\
z_{t+i} & \sim \mathcal{N} (\mu_{t+i}, \sigma_{t+i}),
\end{split}
\label{eq3}
\end{equation}
where $\mathcal{N}(\mu, \sigma)$ is normal distribution with mean value $\mu$ and standard deviation $\sigma$, $\sim$ is the operation of sampling, and $h^E_{t+i-1}$ is the hidden state of the encoder $E$ which contains information from $z_{<t+i-1}$ and ${\widehat{p}_{<t+i-1}}$. 

The decoder $D$ can synthesize pose sequence recurrently according to the joint posterior distribution of $z_{\leq t+i}$ and $\widehat{p}_{<t+i}$ conditioned by $\mathcal{A}$ as
\begin{equation}
(\widehat{p}_{t+i}, h^{D}_{t+i}) = D(\widehat{p}_{t+i-1}, h^{D}_{t+i-1}, z_{t+i} |\mathcal{A}),
\label{eq33}
\end{equation}
where $h^D_{t+i-1}$ is the hidden state of $D$, which involves information from $\widehat{p}_{<t+i-1}$ and $z_{< t+i}$.

\subsubsection{Learning of ACT-VAE}
Suppose the initial input image is denoted as $I_0$ and its pose is $p_0$, ACT-VAE predicts pose sequence for action category $\mathcal{A}$ as $\{\widehat{p}_{i}\}_{i=1,...,N}$ (i.e., $\widehat{p}_{\leq N}$).
Thus, ACT-VAE is to synthesize future pose by optimizing conditional posterior probability $\mathcal{P}_\theta (\widehat{p}_{\leq N}|p_{0}, \mathcal{A})$, which is approximated by a network with parameters $\theta$. Directly computing $\mathcal{P}_\theta (\widehat{p}_{\leq N}|p_{0}, \mathcal{A})$ is intractable since it is difficult to compute its probability density function. In VAE \cite{kingma2013auto}, regarding the probability distribution of $\mathcal{P}_\theta (\widehat{p}_{\leq N}|p_{0}, \mathcal{A})$, we maximize its lower bound instead.
And this lower bound can be obtained with Jensen's inequality as
\begin{equation}
\small
{\rm ln} (\mathcal{P}_\theta (\widehat{p}_{\leq N}|p_{0}, \mathcal{A}) ) \geq \mathbb{E}_{z\sim \mathcal{Q}_\phi(z|\widehat{p},\mathcal{A})} \bigg [ {\rm ln} \frac{\mathcal{P}_\theta (\widehat{p}_{\leq N}, z|p_{0}, \mathcal{A})}{\mathcal{Q}_\phi (z|\widehat{p},\mathcal{A})} \bigg ],\\
\label{vae1}
\end{equation}
where $\mathcal{Q}_\phi(z|\widehat{p}, \mathcal{A})$ is a posterior distribution and ${\rm ln}$ is the operation of computing natural logarithm. We further notice that $\mathcal{P}_\theta (\widehat{p}_{\leq N},z|p_{0}, \mathcal{A})$ and $\mathcal{Q}_\phi(z|\widehat{p}, \mathcal{A})$ can be decomposed as:
\begin{equation}
\small
\begin{split}
{\rm ln}(\mathcal{P}_\theta (\widehat{p}_{\leq N}, z|p_{0}, \mathcal{A}))&= {\rm ln}(\prod_{t=1:N} \mathcal{P}_\theta (\widehat{p}_t|z_{\leq t}, \widehat{p}_{< t},\mathcal{A})\mathcal{P}_\phi (z_t)),\\
{\rm ln}(\mathcal{Q}_\phi(z|\widehat{p}, \mathcal{A}))&={\rm ln} (\prod_{t=1:N} \mathcal{Q}_\phi (z_t|z_{< t}, \widehat{p}_{< t},\mathcal{A})),
\end{split}
\label{vae2}
\end{equation}
where $\mathcal{P}_\theta (\widehat{p}_t|z_{\leq t},\widehat{p}_{<t}, \mathcal{A})$ and $\mathcal{Q}_\phi (z_t|z_{< t}, \widehat{p}_{< t},\mathcal{A})$ are two posterior distributions that are approximated by $D$ and $E$ in ACT-VAE. $\mathcal{P}_\phi (z_t)$ is the prior distribution for $z_t$.
According to Eq.~\eqref{vae1} and Eq.~\eqref{vae2}, we can obtain the lower bound of ${\rm ln} (\mathcal{P}_\theta (\widehat{p}_{\leq N}|p_{0}, \mathcal{A}) )$ as
\begin{equation}
\begin{split}
\mathbb{E}_{z\sim \mathcal{Q}_\phi(z|\widehat{p},\mathcal{A})}\bigg [ \sum_{t=1:N}  (&{\rm ln} (\mathcal{P}_\theta (\widehat{p}_t|z_{\leq t}, \widehat{p}_{<t}, \mathcal{A}))+\\{\rm ln} (\mathcal{P}_{\phi}(z_t))-&{\rm ln} (\mathcal{Q}_\phi (z_t|z_{<t}, \widehat{p}_{<t}, \mathcal{A}))) \bigg ].
\end{split}
\label{eq1}
\end{equation}
Moreover, it is trivial to obtain the expression of an objective to optimize when the initial input pose is denoted as $p_t$, by replacing the corresponding time index.
We will show the superiority of this novel optimization objective for the accuracy of prediction with experiments in Sec.~\ref{sec.exp}. 
Note
\begin{equation}
\begin{split}
&\mathbb{E}_{z\sim \mathcal{Q}_\phi} \bigg [ {\rm ln} (\mathcal{P}_{\phi}(z_t))-{\rm ln} (\mathcal{Q}_\phi (z_t|z_{<t}, \widehat{p}_{<t}, \mathcal{A}))\bigg ] = \\-&\mathbb{E}_{z\sim \mathcal{Q}_\phi}\bigg [{\rm ln} \frac{\mathcal{Q}_\phi (z_t|z_{<t}, \widehat{p}_{<t}, \mathcal{A})}{\mathcal{P}_{\phi}(z_t)} \bigg ],
\end{split}
\label{eq4}
\end{equation}
which is the negative KL-divergence between two distributions of $\mathcal{Q}_\phi (z_t|z_{<t}, \widehat{p}_{<t}, \mathcal{A})$ and $\mathcal{P}_{\phi}(z_t)$. This is the cost function of the encoder in ACT-VAE.

For the decoder, its objective  is $\mathcal{P}_\theta (\widehat{p}_{t}|z_{\leq t}, \widehat{p}_{<t}, \mathcal{A})$ in Eq.~\eqref{eq1}. Maximizing it leads the predicted pose sequence to be close to its ground truth.
It is trivial to obtain the expression of an objective to optimize when the initial input image is denoted as $I_t$, by replacing the corresponding time index.

\subsubsection{Training Objective}
If the input pose of ACT-VAE is $p_t$ and we predict $N$ frames, then the optimization target of ACT-VAE is to minimize the distance $\mathcal{L}_{dis}$ and KL-divergence $\mathcal{L}_{div}$, as
\begin{equation}
\begin{split}
&\mathcal{L}_{dis}= \sum_{t'=t+1}^{t+N}|| \widehat{p}_{t'}-p_{t'}||_1, \\
&\mathcal{L}_{div}=\sum_{t'=t+1}^{t+N} {\rm KL}(\mathcal{N}(\mu_{t'}, \sigma_{t'}), \mathcal{P}_\phi (z_{t'})), \\
&\mathcal{L}_{vae}=\lambda_1 \, \mathcal{L}_{dis} + \lambda_2 \, \mathcal{L}_{div},
\end{split}
\label{eqloss}
\end{equation}
where $\mathcal{L}_{vae}$ is the loss to optimize for ACT-VAE, $\widehat{p}_{t'}$ is the generated pose from ACT-VAE, and $p_{t'}$ is its corresponding ground truth at time $t'$. 
The prior distribution $\mathcal{P}_{\phi}(z_t)$ is assumed to be the standard normal distribution $\mathcal{N}(0, {\rm I})$.
${\rm KL}$ is to compute the KL-divergence between two distributions. $\lambda_1$ and $\lambda_2$ are loss weights that are obtained by using the grid search on the validation set.
Note that the action label $\mathcal{A}$ should be consistent with the input image/pose during training, while can be inconsistent with the input during inference to control which action type to generate.

\subsubsection{Inference}
Given the pose of input image $p_t$ and the target action label $\mathcal{A}$, we aim to generate pose sequence $\{\widehat{p}_{t+i}\}_{i=1,...,N}$ during inference. To obtain $\widehat{p}_{t+i}$, we first sample latent variable $z_{t+i}$ with Eq.~\eqref{eq3} and then use it to compute $\widehat{p}_{t+i}$ with Eq.~\eqref{eq33}. Obviously, the process to sample $\{z_{t+i}\}_{i=1,...,N}$ and generate the pose sequence $\{\widehat{p}_{t+i}\}_{i=1,...,N}$ is same for both training and inference.

Our modeling on latent variables differs from the current VRNN works \cite{castrejon2019improved,denton2018stochastic}: 
VRNN models $z_t$ with the posterior $z_t\sim \mathcal{Q}_{\phi}(z_t|z_{<t}, \widehat{p}_{\leq t})$ during training and the prior $z_t\sim \mathcal{P}_{\phi}(z_t|z_{<t}, \widehat{p}_{< t})$ during inference; ACT-VAE models $z_t$ with $z_t\sim \mathcal{Q}_{\phi}(z_t|z_{<t},  \widehat{p}_{<t})$ for both training and inference, resulting in higher accuracy and diversity as proved in Sec.~\ref{sec.vae_com}.
Besides, ACT-VAE also differs from SVG-FP \cite{denton2018stochastic} that models $z_t$ with a fixed prior during inference.

Moreover, as a general approach, ACT-VAE can synthesize image sequences, by connecting it with a plug-and-play network that maps pose sequences to image sequences. To this, we design an effective Pose-to-Image network.

\subsection{Pose-to-Image Network}
P2I network (denoted as $\mathcal{G}$) predicts a realistic image sequence $\{\widehat{I}_{t+i}\}_{i=1,...,N}$ by taking input of a pose sequence $\{p_{t+i}\}_{i=1,...,N}$ (or $\{\widehat{p}_{t+i}\}_{i=1,...,N}$), a still image $I_t$ with its pose $p_t$, and an action label $\mathcal{A}$.  
We employ the encoder-decoder structure in \cite{zhu2019progressive} as the backbone, with our attention mechanism and conditional batch normalization.

\subsubsection{Foreground Attention} Considering the elusive variance in the background of human videos, directly generating an image sequence tends to yield severe artifacts in the background. For the nearly static background in $I_t$, we exploit foreground-background composition with an attention mechanism. It makes the generator concentrate on foreground synthesis, which is our main focus in this paper. Other background synthesis methods will be our future work. Generally, given $I_t$ and $p_t$ and $p_{t+i}$, the procedure to synthesize the target frame is
\begin{equation}
\begin{split}
&(\bar{I}_{t+i},  M_{t+i}) =\mathcal{G}(I_t, p_t, p_{t+i}), \\
&\widehat{I}_{t+i} =  \bar{I}_{t+i} \odot M_{t+i} + I_t \odot (1-M_{t+i}),
\end{split}
\label{eqim}
\end{equation}
where $\widehat{I}_{t+i}$ is the generated frame, $M_{t+i}$ is a soft mask indicating foreground, and $\odot$ refers to Hadamard product. This procedure is denoted as $\widehat{I}_{t+i}=\mathcal{G_M}(I_t, p_t, p_{t+i})$.

\subsubsection{Action Conditional Batch Normalization} 
We utilize task-related conditions into normalization operations to improve results \cite{perez2018film,clark2019adversarial} through incorporating action conditional batch normalization (ACBN) into $\mathcal{G}$. This design is based on the assumption that statistics of intermediate feature maps in $\mathcal{G}$ for each action category should be distinctive. 
From this perspective, we assign affine transformation parameters $\gamma_{\mathcal{A}}$ and $\tau_{\mathcal{A}}$ for BN operations in the decoder of $\mathcal{G}$ with the condition of the action label $\mathcal{A}$ as
\begin{equation}
\widehat{x}=\gamma_{{\mathcal{A}}} \frac{x-\nu_x}{\sqrt{\rho_x^2+\epsilon}} + \tau_{{\mathcal{A}}}, \; (\gamma_{\mathcal{A}},   \tau_{\mathcal{A}}) = \mathcal{B}(\mathcal{A}),
\end{equation}
where $\nu_x$ and $\rho_x$ are mean and variance computed from input feature map $x$, and $\epsilon$ is a small positive constant for numerical stability. $\gamma_{\mathcal{A}}$ and $\tau_{\mathcal{A}}$ for each BN operation in decoder are predicted by $\mathcal{A}$ from a network $\mathcal{B}$ as shown in Fig.~\ref{fig:framework}. $\mathcal{B}$ is an embedding layer with the input of action labels.

\subsubsection{Training P2I Network}
The training objective of the P2I network consists of reconstruction and adversarial loss. We employ both pixel-level and perceptual-level \cite{johnson2016perceptual,zhu2017unpaired} reconstruction loss as
\begin{equation}
\mathcal{L}_{rec} = \sum_{i=1}^N \sum_{m=0}^5 \mathbb{E}( ||\Phi_m(\widehat{I}_{t+i}) - \Phi_m(I_{t+i})||_1),
\label{eqrec}
\end{equation}
where $\widehat{I}_{t+i}=\mathcal{G_M}(I_t, p_t, p_{t+i})$, $\mathbb{E}$ is the operation to compute mean value, $\Phi_0(\cdot)$ is the raw pixel space, $\Phi_1(\cdot)$ to $\Phi_5(\cdot)$ are five feature spaces of an ImageNet-pretrained VGG-16 network \cite{johnson2016perceptual}.
Further, adversarial learning is proved to be effective in human video synthesis \cite{yang2018pose,wang2018video}. Thus, we use the form of LS-GAN \cite{mao2017least} for adversarial learning, as
\begin{equation}
\begin{aligned}
\mathcal{L}_{gan_d} = &\mathbb{E} ((\mathcal{D}(I_{t+i} \dagger p_{t+i})-1)^2) \\
+ &\mathbb{E} ((\mathcal{D}(\widehat{I}_{t+i} \dagger p_{t+i}))^2), \\
\mathcal{L}_{gan_g} = &\mathbb{E}((\mathcal{D}(\widehat{I}_{t+i} \dagger p_{t+i})-1)^2),
\end{aligned}
\label{gan1}
\end{equation}
where $i \in [1, N]$, $\dagger$ is the operation of channel concatenation, $\mathcal{D}$ is the discriminator and $\mathcal{L}_{gan_d}$ is its loss term, $\mathcal{L}_{gan_g} $  is the loss term for the P2I network.
To stabilize adversarial learning, we utilize feature match loss $\mathcal{L}_{aux}$ \cite{wang2018high} as an auxiliary part of the adversarial loss, which is the distance between real and fake samples in the feature space of $\mathcal{D}$.
Compared with the generation with $\{p_{t+i}\}_{i=1,...,N}$, the generation with $\{\widehat{p}_{t+i}\}_{i=1,...,N}$ is harder since there is no ground truth.
To address this issue, we adopt a term of adversarial loss, similar to Eq.~\eqref{gan1}, as
\begin{equation}
\begin{aligned}
\mathcal{L}_{gan_{\widehat{d}}} &= \mathbb{E} ((\mathcal{D}( \mathcal{G_M}(I_t, p_t, \widehat{p}_{t+i}) \dagger \widehat{p}_{t+i})-0)^2), \\
\mathcal{L}_{gan_{\widehat{g}}} &= \mathbb{E}((\mathcal{D}(\mathcal{G_M}(I_t, p_t, \widehat{p}_{t+i}) \dagger \widehat{p}_{t+i})-1)^2).
\end{aligned}
\label{gan2}
\end{equation}
In summary, the overall loss terms for $\mathcal{G}$ and $\mathcal{D}$ are
\begin{equation}
\begin{aligned}
&\mathcal{L}_{d} = \lambda_3 \, (\mathcal{L}_{gan_d}+\mathcal{L}_{gan_{\widehat{d}}}), \\
&\mathcal{L}_{g} = \lambda_4 \, (\mathcal{L}_{gan_g}+\mathcal{L}_{gan_{\widehat{g}}})  +\lambda_5 \, \mathcal{L}_{aux} +\lambda_6 \, \mathcal{L}_{rec},
\end{aligned}
\label{gan3}
\end{equation}
where $\lambda_3$ to $\lambda_6$ are loss weights that are set based on parameters of pose-guided image generation methods \cite{zhu2019progressive,wang2018high} and grid search. 
$\mathcal{L}_{d}$ is the loss for $\mathcal{D}$ and $\mathcal{L}_{g}$ is the loss for $\mathcal{G}$. 
In our experiments, $\lambda_1=200$, $\lambda_2=0.002$, $\lambda_3=5$, $\lambda_4=5$, $\lambda_5=1$, $\lambda_6=30$.

\begin{algorithm}[t] \caption{Training algorithm for ACT-VAE} \label{alg1}
	\small
	\begin{algorithmic}[1]
		\STATE \textbf{Input:} training image sequences, with their pose sequences and action labels. Initialized ACT-VAE including the encoder $E$ and the decoder $D$;
		\FOR {$i = 0$ to $maxIters$}
		\FOR {each training tuple $(p_t,...,p_{t+N}, \mathcal{A})$ in dataset}
		\STATE Initialize the hidden state of $E$ as $\textbf{0}$ and set it as $h_{\leq t}^E$;
		\STATE Initialize $z_{\leq t} \sim \mathcal{N}(0, I)$;
		\STATE $(\mu_{t+1}, \sigma_{t+1}, h_{t+1}^E) \leftarrow E(p_t, h_{\leq t}^E, z_{\leq t}|\mathcal{A})$;
		\STATE $z_{t+1} \sim \mathcal{N}(\mu_{t+1}, \sigma_{t+1})$;		
		\STATE Initialize hidden state of $D$ as $\textbf{0}$ and set it as $h_{\leq t}^D$;
		\STATE $(\widehat{p}_{t+1}, h_{t+1}^D)\leftarrow D(p_t, h_{\leq t}^D, z_{t+1}|\mathcal{A})$;
		\FOR {$k= 1$ to $N-1$}
		\STATE $(\mu_{t+k+1}, \sigma_{t+k+1}, h_{t+k+1}^E) \leftarrow E(\widehat{p}_{t+k}, h_{t+k}^E, z_{t+k}|\mathcal{A})$;
		\STATE $z_{t+k+1} \sim \mathcal{N}(\mu_{t+k+1}, \sigma_{t+k+1})$;
		\STATE $(\widehat{p}_{t+k+1}, h_{t+k+1}^D)\leftarrow D(\widehat{p}_{t+k}, h_{t+k}^D, z_{t+k+1}|\mathcal{A})$;
		\ENDFOR
		\STATE Compute loss $\mathcal{L}_{vae}$ in Eq.~\eqref{eqloss}, by using $(\widehat{p}_{t+1},...,\widehat{p}_{t+N})$, $(p_{t+1},...,p_{t+N})$, $(z_{t+1},...,z_{t+N})$, $\lambda_1$ and $\lambda_2$;
		\STATE Update the weights of $E$ and $D$;
		\ENDFOR
		\ENDFOR
		\RETURN trained encoder $E$ and decoder $D$ of ACT-VAE;
	\end{algorithmic}
\end{algorithm}

\begin{algorithm}[t] \caption{Training algorithm for P2I network} \label{alg2}
		\small
	\begin{algorithmic}[1]
		\STATE \textbf{Input:} training image sequences, with their pose sequences and action labels.
		\FOR {$i = 0$ to $maxIters$}
		\FOR {each tuple $[(I_t,...,I_{t+N}), (p_t,...,p_{t+N}), \mathcal{A}]$ in dataset}
		\FOR {$k = 1$ to $N$}
		\STATE $\widehat{I}_{t+k} \leftarrow \mathcal{G_M}(I_t, p_t, p_{t+k})$;
		\ENDFOR
		\STATE Compute $\mathcal{L}_{rec}$ by Eq.~\eqref{eqrec};
		\STATE Compute $\mathcal{L}_{gan_g}$ and $\mathcal{L}_{aux}$ by Eq.~\eqref{gan1};
		\STATE Compute $\lambda_4 \, \mathcal{L}_{gan_g}  +\lambda_5 \, \mathcal{L}_{aux} +\lambda_6 \, \mathcal{L}_{rec}$ to update $\mathcal{G}$;
		\STATE Compute $\lambda_3 \, \mathcal{L}_{gan_d}$ by Eq.~\eqref{gan1} to update weights of $\mathcal{D}$;
		\STATE Take $p_t$, $ \mathcal{A}$ as input, ACT-VAE outputs $(\widehat{p}_{t+1},...,\widehat{p}_{t+N})$;
		\FOR {$k = 1$ to $N$}
		\STATE $\widehat{I}_{t+k} \leftarrow \mathcal{G_M}(I_t, p_t, \widehat{p}_{t+k})$;
		\STATE Compute  $\lambda_4 \, \mathcal{L}_{gan_{\widehat{g}}}$ by Eq.~\eqref{gan2} to update weights of $\mathcal{G}$;
		\STATE Compute $\lambda_3 \, \mathcal{L}_{gan_{\widehat{d}}}$ by Eq.~\eqref{gan2} to update weights of $\mathcal{D}$;
		\ENDFOR
		\ENDFOR
		\ENDFOR
		\RETURN trained P2I network $\mathcal{G}$;
	\end{algorithmic}
\end{algorithm}

\begin{algorithm} \caption{Inference algorithm for human action prediction} \label{alg3}
	\small
	\begin{algorithmic}[1]
		\STATE \textbf{Input:} Trained ACT-VAE which includes encoder $E$ and decoder $D$, trained P2I network $\mathcal{G}$, input still image $I_t$ with its pose $p_t$ and action label $\mathcal{A}$;
		\STATE Initialize the hidden state of $E$ as $\textbf{0}$ and set it as $h_{\leq t}^E$;
		\STATE Initialize $z_{\leq t} \sim \mathcal{N}(0, {\rm I})$;
		\STATE $(\mu_{t+1}, \sigma_{t+1}, h_{t+1}^E) \leftarrow E(p_t, h_{\leq t}^E, z_{\leq t}| \mathcal{A})$;
		\STATE $z_{t+1} \sim \mathcal{N}(\mu_{t+1}, \sigma_{t+1})$;
		\STATE Initialize the hidden state of $D$ as $\textbf{0}$ and set it as $h_{\leq t}^D$;
		\STATE $(\widehat{p}_{t+1}, h_{t+1}^D)\leftarrow D(p_t, h_{\leq t}^D, z_{t+1}| \mathcal{A})$;
		\FOR {$k= 1$ to $N-1$}
		\STATE $(\mu_{t+k+1}, \sigma_{t+k+1}, h_{t+k+1}^E) \leftarrow E(\widehat{p}_{t+k}, h_{t+k}^E, z_{t+k}| \mathcal{A})$;
		\STATE $z_{t+k+1} \sim \mathcal{N}(\mu_{t+k+1}, \sigma_{t+k+1})$;
		\STATE $(\widehat{p}_{t+k+1}, h_{t+k+1}^D)\leftarrow D(\widehat{p}_{t+k}, h_{t+k}^D, z_{t+k+1}| \mathcal{A})$;
		\ENDFOR
		\FOR {$k= 1$ to $N$}
		\STATE Compute $\widehat{I}_{t+k}=\mathcal{G_M}(I_t, p_t, \widehat{p}_{t+k})$ by using $\mathcal{A}$ to compute parameters of ACBN for $\mathcal{G}$;
		\ENDFOR
		\RETURN $\{\widehat{I}_{t+k} \}_{k=1,...,N}$ as predicted image sequence;
	\end{algorithmic}
\end{algorithm}

\subsection{Training and Inference Algorithm}
\label{sec2}
To train our framework, ACT-VAE and P2I networks are trained separately.
We denote the operation of sampling as $\sim$, the variable with all zero value as $\textbf{0}$.
The operation $A \leftarrow B$ means that we set the value of $A$ as the output of $B$.
Then our training algorithms are given below.
\begin{itemize}
	\item To train ACT-VAE, we have training data as the pose sequences with their corresponding action labels. A detailed training algorithm for ACT-VAE is shown in Alg.~\ref{alg1}.
	\item To train P2I network, we need image sequences with their pose sequences and action labels. Its training procedure is shown in Alg.~\ref{alg2}.
\end{itemize}
One feature of our ACT-VAE is that the process of sampling latent variables and generating pose sequences are identical in training and inference.
After we have trained the ACT-VAE and P2I networks, we can synthesize image sequence $\{\widehat{I}_{t+k}\}_{k=1,...,N}$ by using Alg.~\ref{alg3}, with the input as a still image $I_t$ with one action label $\mathcal{A}$.

\begin{figure*}[t]
	\begin{center} 
		\includegraphics[width=1.00\linewidth]{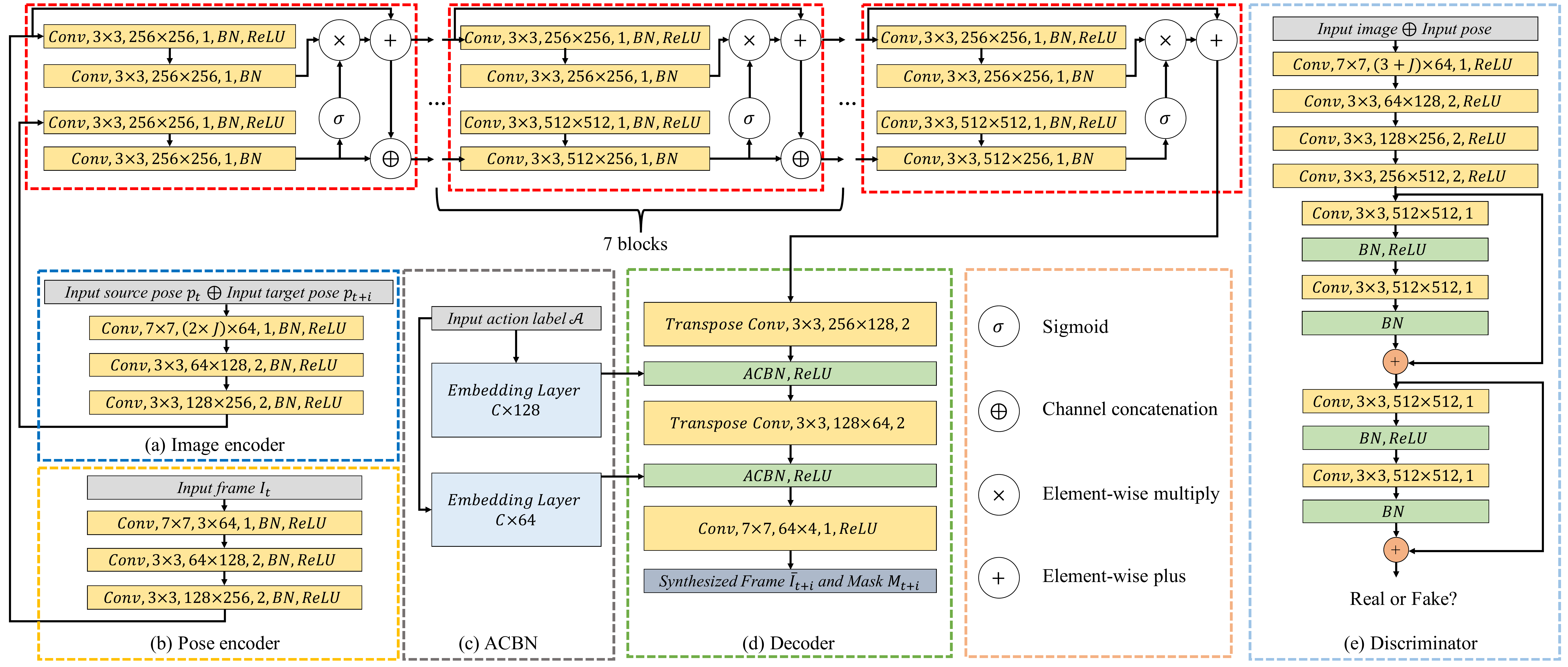}
	\end{center}
	\caption{Detailed structure of the P2I network.}
	\label{fig:structure}
\end{figure*}

\subsection{Network Details}
In experiments, the network configuration for each component in our framework is summarized as the following.
\subsubsection{ACT-VAE}
ACT-VAE is structured like an encoder-decoder, and both encoder and decoder are implemented as one-layer LSTM. The input size of encoder $E$ is $J \times 2 + C + 512$ ($J$ is the joint number in pose representation and $C$ is the number of action categories), the hidden size is 1024, the output size is 512; the input size of decoder $D$ is $J \times 2 + C + 512$, the hidden size is 26, the output size is $J \times 2$.

\subsubsection{P2I Network}
The P2I network has two encoder heads as image encoder $E_I$ (Fig.~\ref{fig:structure} (a)) and pose encoder $E_P$ (Fig.~\ref{fig:structure} (b)), and it contains an image decoder $D_I$ (Fig.~\ref{fig:structure} (d)). Besides, it is trained in adversarial manner, hence it is attached with a discriminator for training (Fig.~\ref{fig:structure} (e)). 
In Fig.~\ref{fig:structure}, ``$Conv, 7\times 7, 3 \times 64, 1, BN, ReLU$'' means that this convolution layer adopts kernel size of $7 \times 7$ with stride size of $1$, and has $3$ input feature channels and $64$ output feature channels. A batch normalization layer and an activation function ReLU is applied to the output of this convolution layer. Meaning of other convolution layers can be interpreted in the same way.

For the synthesis with input as $I_t$, $p_t$ and $p_{t+i}$, the image encoder $E_I$ and pose encoder $E_P$ will transform $I_t$ into image feature, and transform pair $(p_t, p_{t+i})$ into spatial feature. These features are then sent into pose-attentional transfer blocks \cite{zhu2019progressive} (the red rectangles in Fig.~\ref{fig:structure}) and the image decoder $D_I$ to obtain two types of outputs: the produced image and the mask to distinguish foreground and background. To compute the parameters of ACBN in $D_I$, we use two 1D embedding layers $\mathcal{B}$ (Fig.~\ref{fig:structure} (c)) with the input dimension as the number of action categories. 

\begin{figure*}[t]
	\centering
	\newcommand\widthface{0.06}
	\resizebox{.78\linewidth}{!}{
		\begin{tabular}{ccccccccc}
			\includegraphics[align=c,width=\widthface\textwidth]{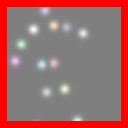} &
			\rotatebox[origin=c]{0}{{w/o $\mathcal{AC}$ I}} &
			\rotatebox[origin=c]{0}{{w/o $\mathcal{AC}$ II}} &
			\rotatebox[origin=c]{0}{{w/o $\mathcal{AZ}$ I}} &
			\rotatebox[origin=c]{0}{{w/o $\mathcal{AZ}$ II}} &
			\rotatebox[origin=c]{0}{{w/o $\mathcal{A}$ I}} &
			\rotatebox[origin=c]{0}{{w/o $\mathcal{A}$ II}} &
			\rotatebox[origin=c]{0}{{Ours I}} &
			\rotatebox[origin=c]{0}{{Ours II}} \\
			
			\includegraphics[align=c,width=\widthface\textwidth]{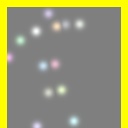} &
			\includegraphics[align=c,width=\widthface\textwidth]{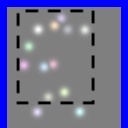} &
			\includegraphics[align=c,width=\widthface\textwidth]{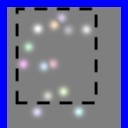} &
			\includegraphics[align=c,width=\widthface\textwidth]{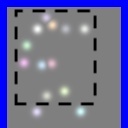} &
			\includegraphics[align=c,width=\widthface\textwidth]{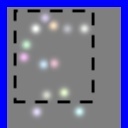} &
			\includegraphics[align=c,width=\widthface\textwidth]{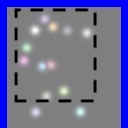} &
			\includegraphics[align=c,width=\widthface\textwidth]{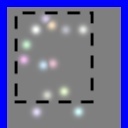} &
			\includegraphics[align=c,width=\widthface\textwidth]{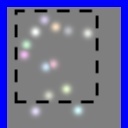} &
			\includegraphics[align=c,width=\widthface\textwidth]{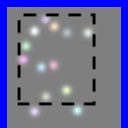} \\

			\includegraphics[align=c,width=\widthface\textwidth]{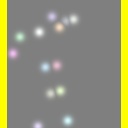} &
			\includegraphics[align=c,width=\widthface\textwidth]{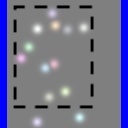} &
			\includegraphics[align=c,width=\widthface\textwidth]{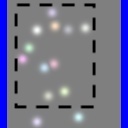} &
			\includegraphics[align=c,width=\widthface\textwidth]{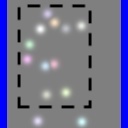} &
			\includegraphics[align=c,width=\widthface\textwidth]{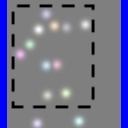} &
			\includegraphics[align=c,width=\widthface\textwidth]{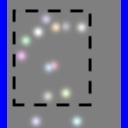} &
			\includegraphics[align=c,width=\widthface\textwidth]{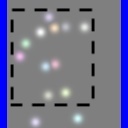} &
			\includegraphics[align=c,width=\widthface\textwidth]{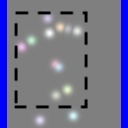} &
			\includegraphics[align=c,width=\widthface\textwidth]{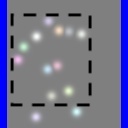} \\
			
			\includegraphics[align=c,width=\widthface\textwidth]{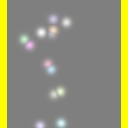} &
			\includegraphics[align=c,width=\widthface\textwidth]{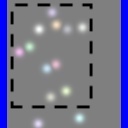} &
			\includegraphics[align=c,width=\widthface\textwidth]{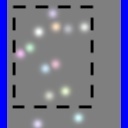} &
			\includegraphics[align=c,width=\widthface\textwidth]{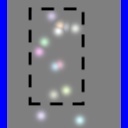} &
			\includegraphics[align=c,width=\widthface\textwidth]{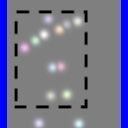} &
			\includegraphics[align=c,width=\widthface\textwidth]{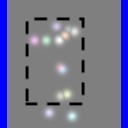} &
			\includegraphics[align=c,width=\widthface\textwidth]{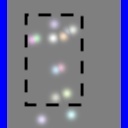} &
			\includegraphics[align=c,width=\widthface\textwidth]{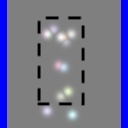} &
			\includegraphics[align=c,width=\widthface\textwidth]{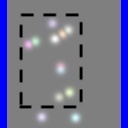} \\
			
			\includegraphics[align=c,width=\widthface\textwidth]{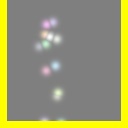} &
			\includegraphics[align=c,width=\widthface\textwidth]{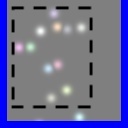} &
			\includegraphics[align=c,width=\widthface\textwidth]{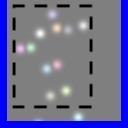} &
			\includegraphics[align=c,width=\widthface\textwidth]{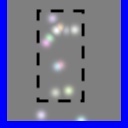} &
			\includegraphics[align=c,width=\widthface\textwidth]{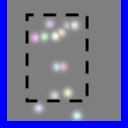} &
			\includegraphics[align=c,width=\widthface\textwidth]{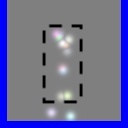} &
			\includegraphics[align=c,width=\widthface\textwidth]{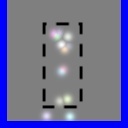} &
			\includegraphics[align=c,width=\widthface\textwidth]{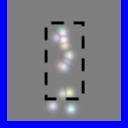} &
			\includegraphics[align=c,width=\widthface\textwidth]{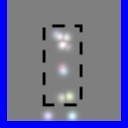} \\
			{real} & \multicolumn{8}{c}{{baseball pitch}}\\

			\includegraphics[align=c,width=\widthface\textwidth]{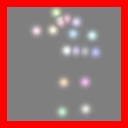} &
			\rotatebox[origin=c]{0}{{w/o $\mathcal{AC}$ I}} &
			\rotatebox[origin=c]{0}{{w/o $\mathcal{AC}$ II}} &
			\rotatebox[origin=c]{0}{{w/o $\mathcal{AZ}$ I}} &
			\rotatebox[origin=c]{0}{{w/o $\mathcal{AZ}$ II}} &
			\rotatebox[origin=c]{0}{{w/o $\mathcal{A}$ I}} &
			\rotatebox[origin=c]{0}{{w/o $\mathcal{A}$ II}} &
			\rotatebox[origin=c]{0}{{Ours I}} &
			\rotatebox[origin=c]{0}{{Ours II}} \\
			\includegraphics[align=c,width=\widthface\textwidth]{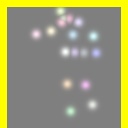} &
			\includegraphics[align=c,width=\widthface\textwidth]{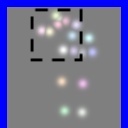} &
			\includegraphics[align=c,width=\widthface\textwidth]{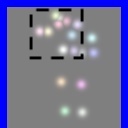} &
			\includegraphics[align=c,width=\widthface\textwidth]{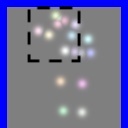} &
			\includegraphics[align=c,width=\widthface\textwidth]{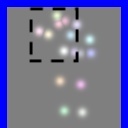} &
			\includegraphics[align=c,width=\widthface\textwidth]{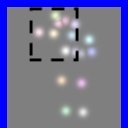} &
			\includegraphics[align=c,width=\widthface\textwidth]{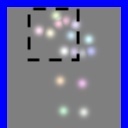} &
			\includegraphics[align=c,width=\widthface\textwidth]{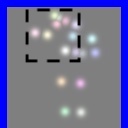} &
			\includegraphics[align=c,width=\widthface\textwidth]{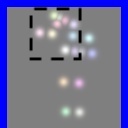} \\
			
			\includegraphics[align=c,width=\widthface\textwidth]{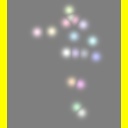} &
			\includegraphics[align=c,width=\widthface\textwidth]{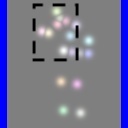} &
			\includegraphics[align=c,width=\widthface\textwidth]{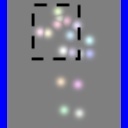} &
			\includegraphics[align=c,width=\widthface\textwidth]{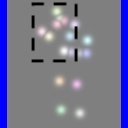} &
			\includegraphics[align=c,width=\widthface\textwidth]{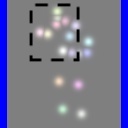} &
			\includegraphics[align=c,width=\widthface\textwidth]{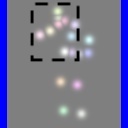} &
			\includegraphics[align=c,width=\widthface\textwidth]{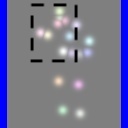} &
			\includegraphics[align=c,width=\widthface\textwidth]{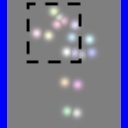} &
			\includegraphics[align=c,width=\widthface\textwidth]{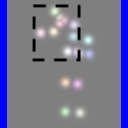} \\
			
			\includegraphics[align=c,width=\widthface\textwidth]{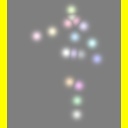} &
			\includegraphics[align=c,width=\widthface\textwidth]{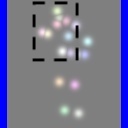} &
			\includegraphics[align=c,width=\widthface\textwidth]{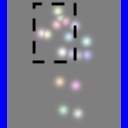} &
			\includegraphics[align=c,width=\widthface\textwidth]{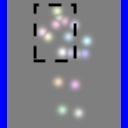} &
			\includegraphics[align=c,width=\widthface\textwidth]{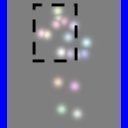} &
			\includegraphics[align=c,width=\widthface\textwidth]{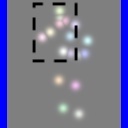} &
			\includegraphics[align=c,width=\widthface\textwidth]{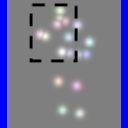} &
			\includegraphics[align=c,width=\widthface\textwidth]{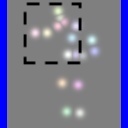} &
			\includegraphics[align=c,width=\widthface\textwidth]{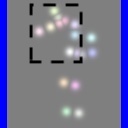} \\
			
			\includegraphics[align=c,width=\widthface\textwidth]{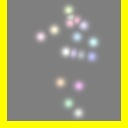} &
			\includegraphics[align=c,width=\widthface\textwidth]{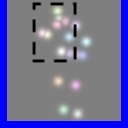} &
			\includegraphics[align=c,width=\widthface\textwidth]{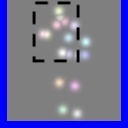} &
			\includegraphics[align=c,width=\widthface\textwidth]{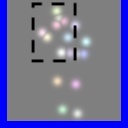} &
			\includegraphics[align=c,width=\widthface\textwidth]{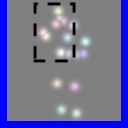} &
			\includegraphics[align=c,width=\widthface\textwidth]{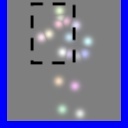} &
			\includegraphics[align=c,width=\widthface\textwidth]{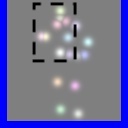} &
			\includegraphics[align=c,width=\widthface\textwidth]{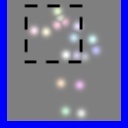} &
			\includegraphics[align=c,width=\widthface\textwidth]{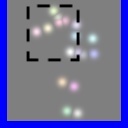} \\
			{real} &\multicolumn{8}{c}{{walking together}}\\
			
	\end{tabular}}
\vspace{-0.1in}
	\caption{Visual comparison for ablation study on key point generation evaluation in the testing set. 
		The column of ``real" is the input image and real future pose sequence.
		The red rectangles are input conditional images, the yellow rectangles are real future pose sequences, and the blue rectangles are the predictions.
		``I" and ``II" are two diverse predictions for identical input, derived from each ablation setting. And the black dotted rectangles are the parts that can notably reflect the accuracy and diversity of our results.	
	} \label{fig:comparison-vae}
\end{figure*}

\begin{table*}[t]
	\centering
	\caption{Quantitative results of ablation study on the key point generation evaluation. ``$L_2$'' is the $L_2$ distance to measure accuracy, and ``Std'' is the standard deviation to measure diversity.}
	\vspace{-0.1in}
    {
		\begin{tabular}{c|c|c|c|c|c|c|c|c}
			\toprule[1pt]
			&\multicolumn{4}{c|}{Penn-action}&\multicolumn{4}{c}{Human3.6M}\\
			\hline
			& w/o $\mathcal{A}$ & w/o $\mathcal{AZ}$&w/o $\mathcal{AC}$ &Ours & w/o $\mathcal{A}$ & w/o $\mathcal{AZ}$&w/o $\mathcal{AC}$&Ours  \\
			\hline
			$L_2$ ($\downarrow$) &29.59 &30.31&34.74&\textbf{28.32}&30.88&31.01&31.37&\textbf{30.41}\\
			\hline
			Std ($\uparrow$) &1.584 &1.462&0.732&\textbf{1.663}&0.730&0.672&0.564&\textbf{0.838}\\
			\bottomrule[1pt]
	\end{tabular}}
	\label{tab:vae_abla}
\end{table*}

\section{Experiments}
\label{sec.exp}
\subsection{Datasets}
To verify our method's generality, we employ datasets containing various action categories, which are Penn-action \cite{zhang2013actemes} and Human3.6M \cite{ionescu2013human3}.
Penn-action contains videos of humans in 15 sport action categories. The total number of videos is 2,326. For each video, 13 human joint annotations are provided as the ground truth of pose.
We adopt the experimental setting of \cite{yunji_neurips_2019} with 9 action categories for experiments, including baseball pitch, baseball swing, clean and jerk, pull ups, golf swing, tennis forehand, tennis serve, jumping jacks and squats.
Besides, we follow the same train/test split of \cite{yunji_neurips_2019} for a fair comparison.

Human3.6M contains various daily human actions, and this dataset provides 17 human joint annotations as the ground truth of pose.
Moreover, to conduct experiments on the action category with obvious motion patterns, we choose 8 action categories from this dataset for experiments: directions, greeting, phoning, posing, purchases, walking, walking dog, and walking together. 
Moreover, we follow the same train/test split of \cite{mao2019learning}.

\textit{This paper focuses on the modeling of human action, thus we experiment on action datasets with static or simple backgrounds to minimize disturbances from backgrounds}. This is the reason why we choose Penn-action and Human3.6M.
The synthesis with dynamic backgrounds and the modeling of general object moving will be our future work.
We set the resolution of both input images and output videos as $128\times 128$, since it is the maximal resolution adopted in existing methods. 

\subsection{Implementation Details}
To train ACT-VAE and P2I networks, we employ Adam optimizer \cite{kingma2014adam} with $\beta_1$ and $\beta_2$ set as 0.5 and 0.999 respectively. The learning rate is $10^{-4}$ and the batch size is 24. 
Our approach is implemented in PyTorch 1.0.1 \cite{paszke2019pytorch}, and runs on an Intel 2.60GHz CPU and TITAN X GPU. 
On average, our framework can create 4 frames in resolution $128\times128$ with a single input image and an action label within 34.93 ms, where 2.57 ms is spent for ACT-VAE and 32.36ms for P2I network.
The model capacity for ACT-VAE and P2I networks are 6.503M and 41.352M respectively.

\begin{figure*}[t]
	\centering
	\newcommand\widthface{0.05}
	\resizebox{.95\linewidth}{!}{
		\begin{tabular}{c@{\hspace{2mm}}c@{\hspace{1mm}}c@{\hspace{1mm}}c@{\hspace{1mm}}c@{\hspace{2mm}}c@{\hspace{2mm}}c@{\hspace{1mm}}c@{\hspace{1mm}}c@{\hspace{1mm}}c}
			{Input}&\multicolumn{4}{c}{{tennis serve}} & {Input}&\multicolumn{4}{c}{{pullup}} \\
			\vspace{-0.05in}
			\includegraphics[width=\widthface\textwidth]{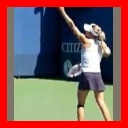} &
			\includegraphics[width=\widthface\textwidth]{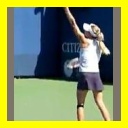} &
			\includegraphics[width=\widthface\textwidth]{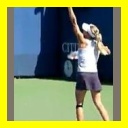} &
			\includegraphics[width=\widthface\textwidth]{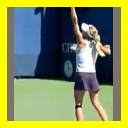} &
			\includegraphics[width=\widthface\textwidth]{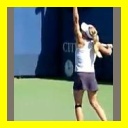} &
			\includegraphics[width=\widthface\textwidth]{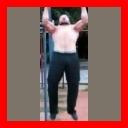} &
			\includegraphics[width=\widthface\textwidth]{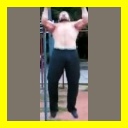} &
			\includegraphics[width=\widthface\textwidth]{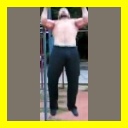} &
			\includegraphics[width=\widthface\textwidth]{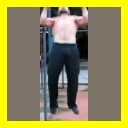}&
			\includegraphics[width=\widthface\textwidth]{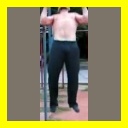} \\
			&&&&&&&&&\\
			
			\rotatebox[origin=c]{0}{w/o mask}&
			\includegraphics[align=c,width=\widthface\textwidth]{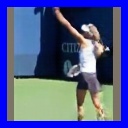} &
			\includegraphics[align=c,width=\widthface\textwidth]{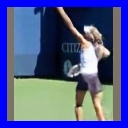} &
			\includegraphics[align=c,width=\widthface\textwidth]{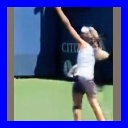} &
			\includegraphics[align=c,width=\widthface\textwidth]{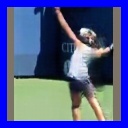} &
			\rotatebox[origin=c]{0}{w/o mask}&
			\includegraphics[align=c,width=\widthface\textwidth]{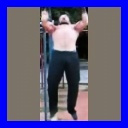} &
			\includegraphics[align=c,width=\widthface\textwidth]{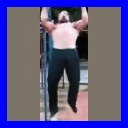} &
			\includegraphics[align=c,width=\widthface\textwidth]{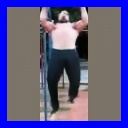}&
			\includegraphics[align=c,width=\widthface\textwidth]{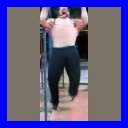}\\
			
			&&&&&&&&&\\
			\rotatebox[origin=c]{0}{w/o ACBN}&
			\includegraphics[align=c,width=\widthface\textwidth]{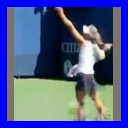} &
			\includegraphics[align=c,width=\widthface\textwidth]{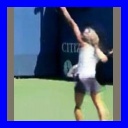} &
			\includegraphics[align=c,width=\widthface\textwidth]{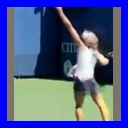} &
			\includegraphics[align=c,width=\widthface\textwidth]{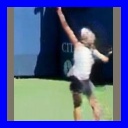} &
			\rotatebox[origin=c]{0}{w/o ACBN}&
			\includegraphics[align=c,width=\widthface\textwidth]{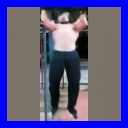} &
			\includegraphics[align=c,width=\widthface\textwidth]{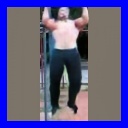} &
			\includegraphics[align=c,width=\widthface\textwidth]{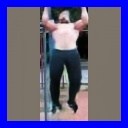}&
			\includegraphics[align=c,width=\widthface\textwidth]{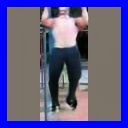}\\
			
			&&&&&&&&&\\
			\rotatebox[origin=c]{0}{Ours}&
			\includegraphics[align=c,width=\widthface\textwidth]{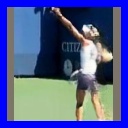} &
			\includegraphics[align=c,width=\widthface\textwidth]{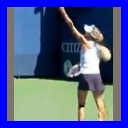} &
			\includegraphics[align=c,width=\widthface\textwidth]{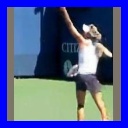} &
			\includegraphics[align=c,width=\widthface\textwidth]{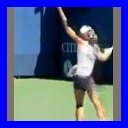} &
			\rotatebox[origin=c]{0}{Ours}&
			\includegraphics[align=c,width=\widthface\textwidth]{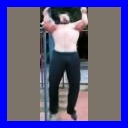} &
			\includegraphics[align=c,width=\widthface\textwidth]{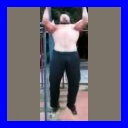} &
			\includegraphics[align=c,width=\widthface\textwidth]{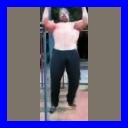}&
			\includegraphics[align=c,width=\widthface\textwidth]{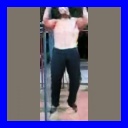}\\
	\end{tabular}}
	\caption{Visual comparison for ablation study on image sequence generation evaluation in the testing set. Input images, real future pose sequences, and the corresponding predictions are marked by red, yellow, and blue rectangles, respectively.
	}
	\label{fig:abla}
\end{figure*}

\subsection{Metrics} 

\subsubsection{Key Points Generation Evaluation} To evaluate the accuracy of estimated future key points, we adopt $L_2$ distance between coordinates of predicted pose sequence and the ground truth, the same as in that of \cite{yang2018pose}. The coordinate range is $[0,127] \times [0,127] $ for $128 \times 128$ images.
Especially, for each sample, we synthesize 100 sequences to compute their distances with the ground truth and report the average value of 10 smallest distances.
Besides, similar to \cite{Prediction-ECCV-2018}, we compute the standard deviation of the predicted key point coordinates from the estimated sequences (these sequences are obtained through the repeated sampling of latent variables for identical input pose) as the indication of how diverse the model predictions are. Specifically, we compute the standard deviation for the coordinate value of each joint, and take the average value of these standard deviations on all joints as the metric. This is also computed with coordinate range as $[0,127] \times [0,127]$. 

\subsubsection{Image Sequence Generation Evaluation} For the evaluation of generated image sequences, we adhere to the protocols in \cite{yunji_neurips_2019}, using FVD \cite{unterthiner2018towards}, $L_2$ distance, action recognition accuracy, and user study.
FVD refers to the Fréchet distance between the deep features of real and generated videos. Such features are gained from the I3D model \cite{carreira2017quo}, as used in \cite{yunji_neurips_2019}.
FVD is set to compare the visual quality, temporal coherence, and diversity for generated videos. 
We also employ metrics utilized in \cite{castrejon2019improved}, which are SSIM \cite{wang2004image} and LPIPS \cite{zhang2018unreasonable}. LPIPS computes perceptual similarity with deep features.

\begin{table*}[t]
	\centering
	\caption{Ablation study on image sequence generation evaluation.}
    {
		\begin{tabular}{c|c|c|c|c|c}
			\toprule[1pt]
			Method  & FVD ($\downarrow$)  & Accuracy ($\uparrow$)  & $L_2$ ($\downarrow$)  & SSIM ($\uparrow$)  & LPIPS ($\downarrow$) \\
			\hline
			w/o mask &1455.2&66.03&50.04 &0.7913 & 0.1132\\
			w/o ACBN &1377.2&68.83&42.16 &0.8028 & 0.1247\\
			Ours &\textbf{1092.8}&\textbf{70.04}& \textbf{39.82}&\textbf{0.8248}&\textbf{0.0908}\\
			\bottomrule[1pt]
	\end{tabular}}
	\label{tab:abla}
\end{table*}

\subsection{Ablation Study}
\subsubsection{Key Point Generation Evaluation} 
\label{abla-key}
We first set ablation studies to explore how extra conditions (action labels) contribute to the prediction in ACT-VAE.
In ACT-VAE, $z_t \sim \mathcal{Q}_\phi (z_t|z_{<t}, \widehat{p}_{<t}, \mathcal{A})$, and we conduct experiments without action labels for the encoder and decoder of ACT-VAE to analyze the role of $\mathcal{A}$.
This setting is called ``w/o $\mathcal{A}$" and $z_t \sim \mathcal{Q}_\phi (z_t|z_{<t}, \widehat{p}_{<t})$.
The corresponding results are shown in Table \ref{tab:vae_abla} (\textit{We use $\downarrow$ to denote ``the lower the better''; use $\uparrow$ to denote ``the higher the better''}). Though its performance is inferior to that with action labels, it is still better than existing methods (as shown in Table \ref{tab:vae}). Thus, the modeling of ACT-VAE does improve pose prediction accuracy even without action labels. 

Moreover, it is inevitable to prove the contribution of the condition $z_{<t}$: 
we build ACT-VAE without action labels, remove the input condition $z_{<t}$, and keep the condition of the past pose for the encoder of ACT-VAE. This setting is called ``w/o $\mathcal{AZ}$", directly sampling $z_t \sim \mathcal{Q}_\phi (z_t|\widehat{p}_{<t})$ like \cite{denton2018stochastic}. Compared with ``w/o $A$" in Table \ref{tab:vae_abla}, ``w/o $AZ$" has worse results for prediction. Thus, the structural novelty of setting $z_{<t}$ as the condition to sample $z_t$ has a great contribution to the accuracy and diversity.

We also verify the significance of temporal coherence by conducting ACT-VAE without action labels and modeling $z_t$ as independent Gaussian distribution.
The results (``w/o $\mathcal{AC}$") are worse than ``w/o $\mathcal{A}$" and ``w/o $\mathcal{AZ}$". Thus, removing temporal coherence reduces performance.
The principle of ``temporal coherence" should be different within various actions, which is called ``individual temporal coherence". The superiority of ``Ours" over ``w/o $\mathcal{A}$" in Table \ref{tab:vae_abla} proves its positive impact. 

In addition, we provide the visual samples for each ablation setting in Fig.~\ref{fig:comparison-vae}, where ``w/o $\mathcal{AC}$ I/II'', ``w/o $\mathcal{AZ}$ I/II'', ``w/o $\mathcal{A}$ I/II'' and ``Ours I/II'' means two diverse predictions for identical input, derived from the ablation setting of ``w/o $\mathcal{AC}$", ``w/o $\mathcal{AZ}$", ``w/o $\mathcal{A}$" and our full setting, respectively. We should note that our full setting leads to the most outstanding visual accuracy and diversity.

\begin{table}[t]
	\centering
	\caption{The results of ACT-VAE and chosen strategies on key point generation evaluation. ``$L_2$'' is the $L_2$ distance to measure accuracy, and ``Std'' is the standard deviation to measure diversity.
	}
	\resizebox{1.0\linewidth}{!}{
		\begin{tabular}{c|c|c|c|c}
			\toprule[1pt]
			\multirow{2}{1.8cm}{Method} & \multicolumn{2}{c|}{Penn-action} & \multicolumn{2}{c}{Human3.6M}\\
			\cline{2-5}
			& $L_2$ ($\downarrow$) & Std ($\uparrow$) & $L_2$ ($\downarrow$) & Std ($\uparrow$) \\
			\hline
			VAE of \cite{yunji_neurips_2019} &32.88&0.895&32.65&0.336\\
			VRNN~\cite{castrejon2019improved} &30.24&1.495&30.96&0.552\\
			SVG-FP~\cite{denton2018stochastic} &31.44&1.543&32.88&0.720\\
			SVG-LP~\cite{denton2018stochastic} &30.72&1.519&31.92&0.648\\
			Dlow~\cite{yuan2020dlow} &55.84&0.617&31.74&\textbf{1.082}\\
			Mix-and-Match~\cite{aliakbarian2020stochastic} &32.14&1.541&34.03&0.772\\
			MT-VAE~\cite{yan2018mt} &31.85&1.328&33.96&0.514\\
			LSTM of ~\cite{zhao2018learning} &33.43&0.000&38.22&0.000\\
			Traj~\cite{mao2019learning} &31.18&0.000&38.55&0.000 \\
			Rep~\cite{mao2020history} &29.03&0.000&32.35&0.000\\
			\hline
			ACT-VAE (w/o $\mathcal{A}$) &29.59&1.584&30.88&0.730\\
			ACT-VAE &\textbf{28.32}&\textbf{1.663}&\textbf{30.41}&0.838\\
			\bottomrule[1pt]
	\end{tabular}}
	\label{tab:vae}
\end{table}

\begin{figure*}[t]
	\centering
	\large
	\newcommand\widthface{0.12}
	\resizebox{0.9\linewidth}{!}{
		\begin{tabular}{cccccccc}
			\includegraphics[align=c,width=\widthface\textwidth]{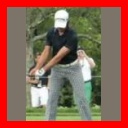} &
			\rotatebox[origin=c]{0}{HL-VP \cite{wichers2018hierarchical}}&
			\rotatebox[origin=c]{0}{LG-VP \cite{villegas2017learning}}&
			\rotatebox[origin=c]{0}{LF-VP \cite{zhao2018learning}}&
			\rotatebox[origin=c]{0}{IVRNN \cite{castrejon2019improved}}&
			\rotatebox[origin=c]{0}{KL-VP \cite{yunji_neurips_2019}}&
			\rotatebox[origin=c]{0}{{KL-VP-Ours}}&
			\rotatebox[origin=c]{0}{Ours}\\
			
			\includegraphics[align=c,width=\widthface\textwidth]{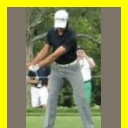} &
			\includegraphics[align=c,width=\widthface\textwidth]{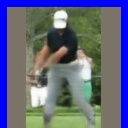} &
			\includegraphics[align=c,width=\widthface\textwidth]{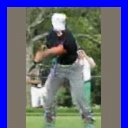} &
			\includegraphics[align=c,width=\widthface\textwidth]{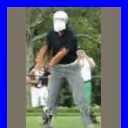} &
			\includegraphics[align=c,width=\widthface\textwidth]{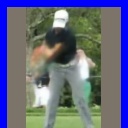} &
			\includegraphics[align=c,width=\widthface\textwidth]{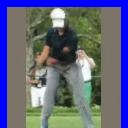} &
			\includegraphics[align=c,width=\widthface\textwidth]{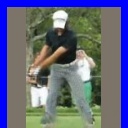} &
			\includegraphics[align=c,width=\widthface\textwidth]{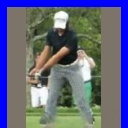} \\
			
			\includegraphics[align=c,width=\widthface\textwidth]{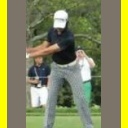} &
			\includegraphics[align=c,width=\widthface\textwidth]{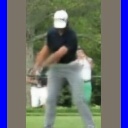} &
			\includegraphics[align=c,width=\widthface\textwidth]{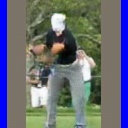} &
			\includegraphics[align=c,width=\widthface\textwidth]{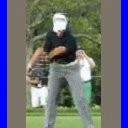} &
			\includegraphics[align=c,width=\widthface\textwidth]{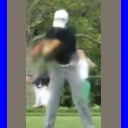} &
			\includegraphics[align=c,width=\widthface\textwidth]{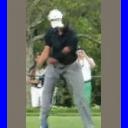} &
			\includegraphics[align=c,width=\widthface\textwidth]{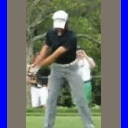} &
			\includegraphics[align=c,width=\widthface\textwidth]{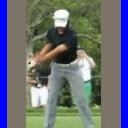} \\

			\includegraphics[align=c,width=\widthface\textwidth]{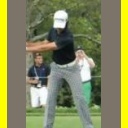} &
			\includegraphics[align=c,width=\widthface\textwidth]{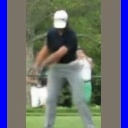} &
			\includegraphics[align=c,width=\widthface\textwidth]{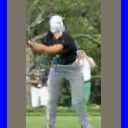} &
			\includegraphics[align=c,width=\widthface\textwidth]{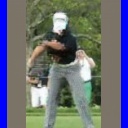} &
			\includegraphics[align=c,width=\widthface\textwidth]{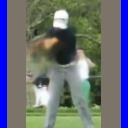} &
			\includegraphics[align=c,width=\widthface\textwidth]{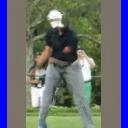} &
			\includegraphics[align=c,width=\widthface\textwidth]{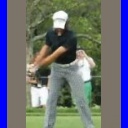} &
			\includegraphics[align=c,width=\widthface\textwidth]{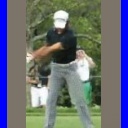} \\
			
			\includegraphics[align=c,width=\widthface\textwidth]{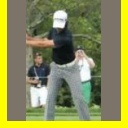} &
			\includegraphics[align=c,width=\widthface\textwidth]{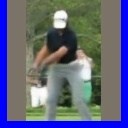} &
			\includegraphics[align=c,width=\widthface\textwidth]{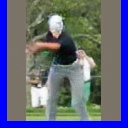} &
			\includegraphics[align=c,width=\widthface\textwidth]{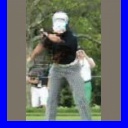} &
			\includegraphics[align=c,width=\widthface\textwidth]{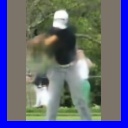} &
			\includegraphics[align=c,width=\widthface\textwidth]{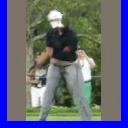} &
			\includegraphics[align=c,width=\widthface\textwidth]{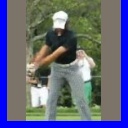} &
			\includegraphics[align=c,width=\widthface\textwidth]{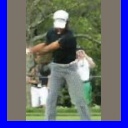} \\
			\LARGE{real} &\multicolumn{7}{c}{\LARGE{golf swing}}\\
			
			\includegraphics[align=c,width=\widthface\textwidth]{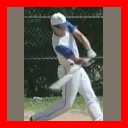} &
			\rotatebox[origin=c]{0}{HL-VP \cite{wichers2018hierarchical}}&
			\rotatebox[origin=c]{0}{LG-VP \cite{villegas2017learning}}&
			\rotatebox[origin=c]{0}{LF-VP \cite{zhao2018learning}}&
			\rotatebox[origin=c]{0}{IVRNN \cite{castrejon2019improved}}&
			\rotatebox[origin=c]{0}{KL-VP \cite{yunji_neurips_2019}}&
			\rotatebox[origin=c]{0}{{KL-VP-Ours}}&
			\rotatebox[origin=c]{0}{Ours}\\
			
			\includegraphics[align=c,width=\widthface\textwidth]{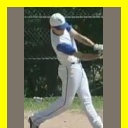} &
			\includegraphics[align=c,width=\widthface\textwidth]{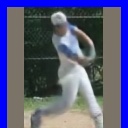} &
			\includegraphics[align=c,width=\widthface\textwidth]{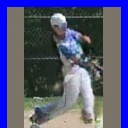} &
			\includegraphics[align=c,width=\widthface\textwidth]{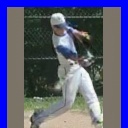} &
			\includegraphics[align=c,width=\widthface\textwidth]{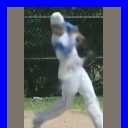} &
			\includegraphics[align=c,width=\widthface\textwidth]{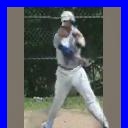} &
			\includegraphics[align=c,width=\widthface\textwidth]{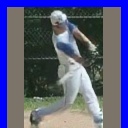} &
			\includegraphics[align=c,width=\widthface\textwidth]{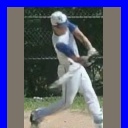} \\
			
			\includegraphics[align=c,width=\widthface\textwidth]{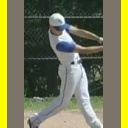} &
			\includegraphics[align=c,width=\widthface\textwidth]{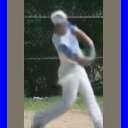} &
			\includegraphics[align=c,width=\widthface\textwidth]{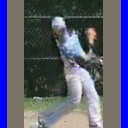} &
			\includegraphics[align=c,width=\widthface\textwidth]{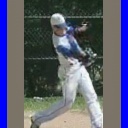} &
			\includegraphics[align=c,width=\widthface\textwidth]{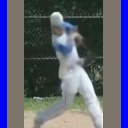} &
			\includegraphics[align=c,width=\widthface\textwidth]{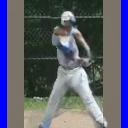} &
			\includegraphics[align=c,width=\widthface\textwidth]{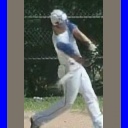} &
			\includegraphics[align=c,width=\widthface\textwidth]{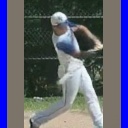} \\
			
			\includegraphics[align=c,width=\widthface\textwidth]{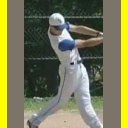} &
			\includegraphics[align=c,width=\widthface\textwidth]{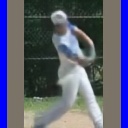} &
			\includegraphics[align=c,width=\widthface\textwidth]{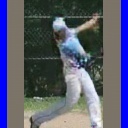} &
			\includegraphics[align=c,width=\widthface\textwidth]{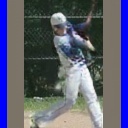} &
			\includegraphics[align=c,width=\widthface\textwidth]{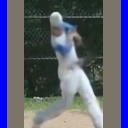} &
			\includegraphics[align=c,width=\widthface\textwidth]{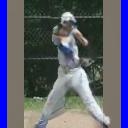} &
			\includegraphics[align=c,width=\widthface\textwidth]{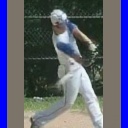} &
			\includegraphics[align=c,width=\widthface\textwidth]{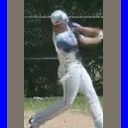} \\
			
			\includegraphics[align=c,width=\widthface\textwidth]{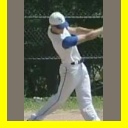} &
			\includegraphics[align=c,width=\widthface\textwidth]{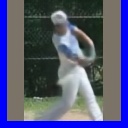} &
			\includegraphics[align=c,width=\widthface\textwidth]{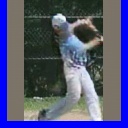} &
			\includegraphics[align=c,width=\widthface\textwidth]{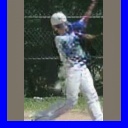} &
			\includegraphics[align=c,width=\widthface\textwidth]{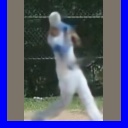} &
			\includegraphics[align=c,width=\widthface\textwidth]{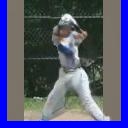} &
			\includegraphics[align=c,width=\widthface\textwidth]{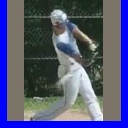} &
			\includegraphics[align=c,width=\widthface\textwidth]{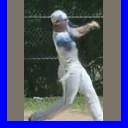} \\
			
			\LARGE{real} & \multicolumn{7}{c}{\LARGE{baseball swing}}\\
			
	\end{tabular}}
	\caption{Visual comparison between our framework and all baselines in the testing set. 
		In the column of ``real", the first row is the input image and the following rows are real future frames.
		The red rectangles are input conditional images, the yellow rectangles are real future frames, and the blue rectangles are the predicted frames from different methods.
	} \label{fig:comparison}
\end{figure*}

\subsubsection{Image Sequence Generation Evaluation} 
\label{sec.abla}
There are two significant parts in our P2I network, i.e., the foreground-background composition strategy and ACBN.
We conduct an ablation study to illustrate their respective importance by deleting them from our framework respectively.
When removing the foreground-background composition strategy, the P2I network directly synthesizes images without mask prediction. We call this setting ``w/o mask''. Removing ACBN from the P2I network, and replacing it with normal BN is denoted as ``w/o ACBN''.
The quantitative results are reported in Table \ref{tab:abla}, and the qualitative samples are shown in Fig.~\ref{fig:abla}. Clearly, removing any of them leads to the degeneration of performance. 

\begin{table*}[t]
	\caption{Comparison between our framework and other methods. Metrics include FVD \cite{unterthiner2018towards}, SSIM \cite{wang2004image}, LPIPS \cite{zhang2018unreasonable}, recognition accuracy (Acc), $L_2$ distance ($L_2$), and the results of ranking from user study (written in the form of ``mean $\pm$ standard deviation'').}
	\vspace{-0.1in}
    {
		\begin{tabular}{c|c|c|c|c|c|c}
			\toprule[1pt]
			Method & FVD ($\downarrow$) & Acc ($\uparrow$) & $L_2$ ($\downarrow$) & SSIM ($\uparrow$)& LPIPS ($\downarrow$)& Ranking ($\downarrow$) \\
			\hline
			KL-VP \cite{yunji_neurips_2019}& 1431.2&67.88 & 44.47&0.7929&0.1127&3.225 $\pm$ 0.352 \\
			LG-VP \cite{villegas2017learning}&  1982.1&52.18 &61.39&0.7433&0.1428&6.150 $\pm$ 0.522 \\
			HL-VP \cite{wichers2018hierarchical}&  2814.3&45.21&60.92&0.7231&0.1459&6.300 $\pm$ 0.143 \\
			LF-VP \cite{zhao2018learning}&  1736.9&56.85&58.54&0.7452&0.1368& 4.290 $\pm$ 0.850\\
			IVRNN \cite{castrejon2019improved}& 1970.2&67.35&46.30&0.7548&0.1297&3.590 $\pm$ 0.892 \\
			KL-VP-Ours&  1318.9 & 62.33&48.61&0.7677&0.1221&2.955 $\pm$ 0.241 \\
			Ours&  \textbf{1092.8}&  \textbf{70.04}& \textbf{39.82}&\textbf{0.8248}&\textbf{0.0908}&\textbf{1.490} $\pm$ 0.292 \\
			\bottomrule[1pt]
	\end{tabular}}
	\label{tab:fvd}
\end{table*}

\subsection{Comparison with Existing Methods}

\subsubsection{Baselines}
\label{baseline-sec}

\noindent For the keypoint generation evaluation, we choose current state-of-the-art strategies which include the pose generation module of KL-VP~\cite{yunji_neurips_2019}, IVRNN~\cite{castrejon2019improved}, SVG-FP~\cite{denton2018stochastic}, SVG-LP~\cite{denton2018stochastic}, Dlow~\cite{yuan2020dlow}, MT-VAE~\cite{yan2018mt}, pose prediction network of LF-VP~\cite{zhao2018learning}, Traj~\cite{mao2019learning} and Rep~\cite{mao2020history}.
KL-VP, MT-VAE and Dlow utilize VAE structures, IVRNN and SVG-LP employ architectures of VRNN, LF-VP and Traj and Rep are all deterministic models for prediction.
For fairness, we change the input of all chosen approaches to our pose representation and concatenate it with the action label, and retrain their models with their released codes for comparison. Besides, we align their settings with ours, including training/testing split and training epochs.

\noindent Moreover, five representative methods are taken as baselines for image sequence generation evaluation, including HL-VP~\cite{wichers2018hierarchical}, LG-VP~\cite{villegas2017learning}, LF-VP~\cite{zhao2018learning}, KL-VP~\cite{yunji_neurips_2019} and IVRNN~\cite{castrejon2019improved}.
HL-VP \cite{wichers2018hierarchical} is a typical GAN-based approach. LG-VP \cite{villegas2017learning}, LF-VP \cite{zhao2018learning} and KL-VP \cite{yunji_neurips_2019} all produce videos by first predicting pose sequences. IVRNN \cite{castrejon2019improved} achieves the best results among current works adopting VRNN.
We use the authors' released codes and unify the training/testing setting for fairness. 
Further, the comparison is conducted on Penn-action following the setting of \cite{yunji_neurips_2019}.
Since our training setting is the same as that of \cite{yunji_neurips_2019}, we use its pre-trained model for comparison.
Moreover, we retrain models of \cite{zhao2018learning,villegas2017learning,wichers2018hierarchical,castrejon2019improved} with our task setting for evaluation.

\subsubsection{Key Point Generation Evaluation}
\label{sec.vae_com}
To illustrate the superiority of ACT-VAE in future pose sequence prediction, we compare ACT-VAE with the VAE proposed by \cite{yunji_neurips_2019}, the hierarchical VRNN designed by \cite{castrejon2019improved}, the straightforward LSTM network adopted in \cite{zhao2018learning}, SVG-FP and SVG-LP \cite{denton2018stochastic}, MT-VAE~\cite{yan2018mt}, Dlow~\cite{yuan2020dlow}, Mix-and-Match~\cite{aliakbarian2020stochastic}, Traj~\cite{mao2019learning} and Rep~\cite{mao2020history}.
For fairness, we change the input of all chosen baselines to our pose representation and concatenate it with the action label.
Significantly, all methods are implemented with their public source codes.
As listed in Table \ref{tab:vae}, ACT-VAE yields the lowest $L_2$ distance.
As for diversity, ACT-VAE achieves higher standard deviations compared with most of the baselines.
Although Dlow results in greater diversity on Human3.6M, ACT-VAE has lower $L_2$ errors. Besides, ACT-VAE has superior results than Dlow on Penn-action in terms of accuracy and diversity.
It is mainly the higher complexity of motion patterns in Penn-action over Human3.6M, which causes poor results of Dlow on Penn-action.
Thus, ACT-VAE is of higher accuracy and diversity than these approaches on the whole.

\subsubsection{Image Sequence Generation Evaluation}
\noindent \textbf{Quantitative results.}
We unify the training and testing settings of all methods, and the experiments are conducted on the Penn-action dataset.
The comparison of FVD for different approaches is given in Table \ref{tab:fvd}. ``KL-VP-Ours'' is produced by synthesizing pose sequences with the VAE structure of \cite{yunji_neurips_2019} and using our P2I network to obtain image sequences. This table shows that our framework yields the lowest FVD. 
Next, as in \cite{yunji_neurips_2019}, we also use action recognition accuracy to evaluate the plausibility of synthesized videos. To this end, we train a network for action recognition with the structure of two-stream CNN \cite{simonyan2014two} on the Penn-Action dataset, which achieves accuracy 82.33\% on real testing videos. And it is clearly in Table \ref{tab:fvd} that the recognition accuracy on our synthesized results is higher than others. It proves that our synthesized motion is in accordance with the ground truth. Further, we compute the $L_2$ distance in pixel-level between the synthesized image sequence and the ground truth, and a lower $L_2$ distance suggests better performance. The results in Table \ref{tab:fvd} show that the $L_2$ distance between our prediction and the ground truth is the lowest. Moreover, it is also evident in Table \ref{tab:fvd} that our approach has the highest SSIM while the lowest LPIPS. This outcome further illustrates our superiority.

Finally, a user study is conducted to check the visual quality of generation following the strategy of \cite{yunji_neurips_2019}. For each question, there is a real video for reference and seven synthesized videos by diverse strategies. The order of these seven videos is randomly chosen, and we ask users to rank them based on quality and accuracy of prediction.
We invite 20 participants using Google Form, and report the average ranking for each method. 
Each participant is required to answer 30 questions.
Results reported in Table \ref{tab:fvd} clearly show that the ranking of our results is the highest with low variance. Thus, our results are the best in human perception compared with these baselines.

\noindent \textbf{Qualitative results.} Video prediction results of our approach and the baselines on several categories of action in Penn-Action, are shown in
Fig.~\ref{fig:comparison}. Our synthesized videos give both the realistic image sequences and the plausible motion for the input action condition.
It is also clear that, compared with these baselines, our method achieves improvements in both the visual and the dynamics quality.
We note KL-VP and KL-VP-Ours are the strongest baselines, while our results are of higher visual realism. 
For \cite{zhao2018learning,villegas2017learning}, the quality of appearance and accuracy of motion prediction have more room for improvement.
And our results are also more dynamic and sharper than those of \cite{wichers2018hierarchical} and  \cite{castrejon2019improved}.

\begin{figure*}[t]
	\centering
	\newcommand\widthface{0.12}
	\begin{tabular}{c@{\hspace{2mm}}c@{\hspace{2mm}}c@{\hspace{2mm}}c@{\hspace{2mm}}c@{\hspace{2mm}}c@{\hspace{2mm}}c}
		{Input}&{\cite{yunji_neurips_2019}}&{Ours}&{\cite{yunji_neurips_2019}}&{Ours}&{\cite{yunji_neurips_2019}}&{Ours}\\
		\includegraphics[align=c, width=\widthface\textwidth]{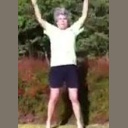} &
		\includegraphics[align=c, width=\widthface\textwidth]{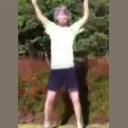} &
		\includegraphics[align=c, width=\widthface\textwidth]{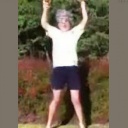} &
		\includegraphics[align=c, width=\widthface\textwidth]{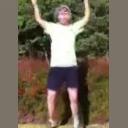} &
		\includegraphics[align=c, width=\widthface\textwidth]{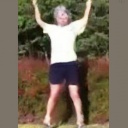} &
		\includegraphics[align=c, width=\widthface\textwidth]{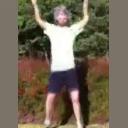} &
		\includegraphics[align=c, width=\widthface\textwidth]{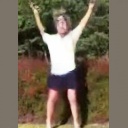} \\
		
		&
		\includegraphics[align=c, width=\widthface\textwidth]{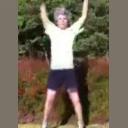} &
		\includegraphics[align=c, width=\widthface\textwidth]{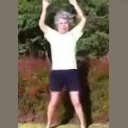} &
		\includegraphics[align=c, width=\widthface\textwidth]{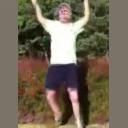} &
		\includegraphics[align=c, width=\widthface\textwidth]{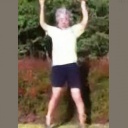} &
		\includegraphics[align=c, width=\widthface\textwidth]{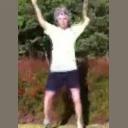} &
		\includegraphics[align=c, width=\widthface\textwidth]{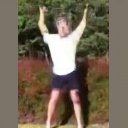} \\
		
		&
		\includegraphics[align=c, width=\widthface\textwidth]{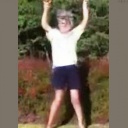} &
		\includegraphics[align=c, width=\widthface\textwidth]{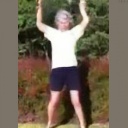} &
		\includegraphics[align=c, width=\widthface\textwidth]{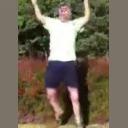} &
		\includegraphics[align=c, width=\widthface\textwidth]{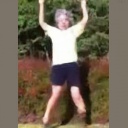} &
		\includegraphics[align=c, width=\widthface\textwidth]{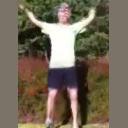} &
		\includegraphics[align=c, width=\widthface\textwidth]{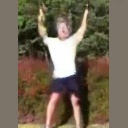} \\
		
		&
		\includegraphics[align=c, width=\widthface\textwidth]{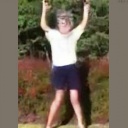} &
		\includegraphics[align=c, width=\widthface\textwidth]{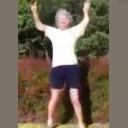} &
		\includegraphics[align=c, width=\widthface\textwidth]{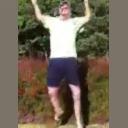} &
		\includegraphics[align=c, width=\widthface\textwidth]{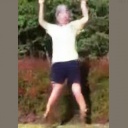} &
		\includegraphics[align=c, width=\widthface\textwidth]{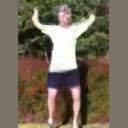} &
		\includegraphics[align=c, width=\widthface\textwidth]{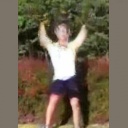} \\
		
		& \multicolumn{2}{c}{$\underbrace{\quad \quad \quad \quad \quad \quad \quad \quad}$} & \multicolumn{2}{c}{$\underbrace{\quad \quad \quad \quad \quad \quad \quad \quad}$} & \multicolumn{2}{c}{$\underbrace{\quad \quad \quad \quad \quad \quad \quad \quad}$} \\
		
		&\multicolumn{2}{c}{ {$\mathcal{A}_1$ jumping jacks}} & \multicolumn{2}{c}{ {$\mathcal{A}_2$ pull up}} & \multicolumn{2}{c}{ {$\mathcal{A}_3$ clean and jerk}}\\
	\end{tabular}
	\caption{our framework can control action types of synthesized image sequences, via controlling the input of action labels.
	}
	\label{fig:control}
\end{figure*}

 \begin{table}[t]
	\centering
	\caption{User preference comparison in the user study. ``Ours" is the percentage that our result is preferred, ``Other" is the percentage that other method is preferred, ``Same" is the percentage that the user has no preference.}
	\label{evaluation2}
     {
		\begin{tabular}{c|ccc}
			\toprule
			Methods & Other & Same & Ours\\
			\midrule
			VAE of \cite{yunji_neurips_2019}  & 15.5\% & 8.5\% &76.0\% \\
			VRNN~\cite{castrejon2019improved}& 24.0\% & 4.5\% & 71.5\% \\
			SVG-FP~\cite{denton2018stochastic} & 20.5\% & 6.0\% & 73.5\% \\
			SVG-LP~\cite{denton2018stochastic} & 18.0\% & 13.5\% & 68.5\% \\
			Dlow~\cite{yuan2020dlow}  & 20.5\% & 12.5\% & 67.0\% \\
			Mix-and-Match~\cite{aliakbarian2020stochastic} & 17.5\% & 10.5\% & 72.0\%\\
			MT-VAE~\cite{yan2018mt}  & 8.0\% &  17.5\% & 74.5\% \\
			LSTM of ~\cite{zhao2018learning} & 12.5\% &  5.5\% & 82.0\% \\
			Traj~\cite{mao2019learning}  & 19.5\% &  10.0\% &70.5\% \\
			Rep~\cite{mao2020history}  & 9.0\% &  8.5\% & 82.5\% \\
			\bottomrule
	\end{tabular}}
\end{table}

\begin{figure*}[t]
	\centering
	\newcommand\widthface{0.09}
	\large
	\resizebox{1.0\linewidth}{!}{
		\begin{tabular}{c@{\hspace{1mm}}c@{\hspace{1mm}}c@{\hspace{1mm}}c@{\hspace{1mm}}c@{\hspace{5mm}}c@{\hspace{1mm}}c@{\hspace{1mm}}c@{\hspace{1mm}}c@{\hspace{1mm}}c}
			\multicolumn{5}{c}{baseball pitch}&\multicolumn{5}{c}{golf swing}\\
			\includegraphics[align=c,width=\widthface\textwidth]{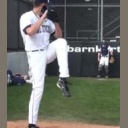} &
			\includegraphics[align=c,width=\widthface\textwidth]{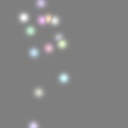} &
			\includegraphics[align=c,width=\widthface\textwidth]{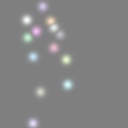} &
			\includegraphics[align=c,width=\widthface\textwidth]{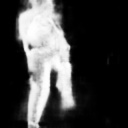} &
			\includegraphics[align=c,width=\widthface\textwidth]{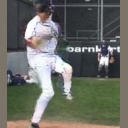} &
			\includegraphics[align=c,width=\widthface\textwidth]{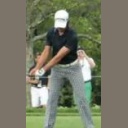} &
			\includegraphics[align=c,width=\widthface\textwidth]{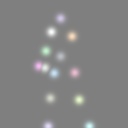} &
			\includegraphics[align=c,width=\widthface\textwidth]{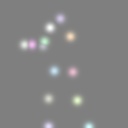} &
			\includegraphics[align=c,width=\widthface\textwidth]{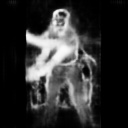} &
			\includegraphics[align=c,width=\widthface\textwidth]{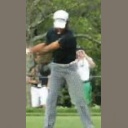} \\
			\specialrule{0em}{5pt}{5pt}
			\multicolumn{5}{c}{pullup}&\multicolumn{5}{c}{tennis serve}\\
			\includegraphics[align=c,width=\widthface\textwidth]{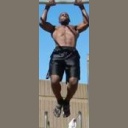} &
			\includegraphics[align=c,width=\widthface\textwidth]{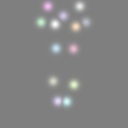} &
			\includegraphics[align=c,width=\widthface\textwidth]{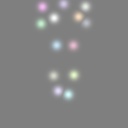} &
			\includegraphics[align=c,width=\widthface\textwidth]{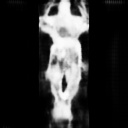} &
			\includegraphics[align=c,width=\widthface\textwidth]{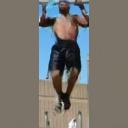} &
			\includegraphics[align=c,width=\widthface\textwidth]{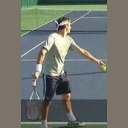} &
			\includegraphics[align=c,width=\widthface\textwidth]{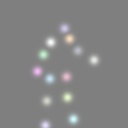} &
			\includegraphics[align=c,width=\widthface\textwidth]{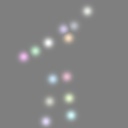} &
			\includegraphics[align=c,width=\widthface\textwidth]{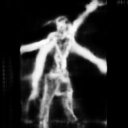} &
			\includegraphics[align=c,width=\widthface\textwidth]{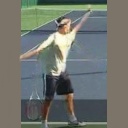} \\
			
			\specialrule{0em}{5pt}{5pt}
			$I_{t}$&$p_t$ & $\widehat{p}_{t+i}$&$M_{t+i}$ &$\widehat{I}_{t+i}$ & $I_{t}$&$p_t$ & $\widehat{p}_{t+i}$&$M_{t+i}$ &$\widehat{I}_{t+i}$\\
	\end{tabular}}
	\caption{The visualization of the mask output from the P2I network.} \label{fig:mask}
\end{figure*}

\subsection{The Control of Action Types in Synthesis}
\label{control}
Our method can control the action types of the synthesized sequences like \cite{yunji_neurips_2019}.
Given an input image, we can provide a list of action labels $\mathcal{A}_1, ..., \mathcal{A}_h$, to synthesize a list of videos whose action types are consistent with the input action labels $\mathcal{A}_1, ..., \mathcal{A}_h$. 
To verify this, we conduct a quantitative analysis as follows. We synthesize all types of action sequences from each testing image via providing various action labels, and report the value of FVD and recognition accuracy for the synthesized image sequences.
The FVD and recognition accuracy of our method are 1257.6 and 58.87; the results of \cite{yunji_neurips_2019} are 1573.7 and 50.04. 
These results demonstrate that our approach can achieve the control of action types in the synthesis and outperforms the strong baseline  \cite{yunji_neurips_2019}.
The qualitative samples for comparison are displayed in Fig.~\ref{fig:control}.

A user study with AB-test is conducted to evaluate the control of action types in synthesis: we invite 20 participants to sees two videos that are synthesized by our method and \cite{yunji_neurips_2019}, and they will choose which one is more consistent with the input action label. 
To demonstrate the performance of controlling action types, the action labels are inconsistent with the input conditional images for synthesis.
Each participant is required to complete 30 pairs of AB-test and 83.7\% of them prefer our method's results.

They are also invited to complete the AB-test for the evaluation of pose sequence: we synthesize pose sequences with our ACT-VAE and all other baselines whose implementations are reported in Sec.~\ref{baseline-sec}. All baselines can synthesize different types of the action sequences since their inputs include the action labels.
To demonstrate the performance of controlling action types, the action labels are inconsistent with the input conditional images for synthesis.
We invite 20 participants to sees two pose sequences that are synthesized by our ACT-VAE and one of the other baselines, and they will choose which one is more consistent with the input action label (they can also choose that they have no preference). 
Each participant is required to complete 100 pairs of AB-test (10 baselines and the comparison with each baseline contains 10 questions). The results are shown in Table \ref{evaluation2}.
These results demonstrate that our ACT-VAE can better implement the control of action types in the synthesis.

\begin{table}[t]
	\centering
	\caption{Results of ACT-VAE with different $\lambda_1$ and $\lambda_2$.}
	\Large
	\resizebox{1.0\linewidth}{!}
	{
		\begin{tabular*}{14.12cm}{p{1.5cm}<{\centering}|p{2cm}<{\centering}|p{2cm}<{\centering}|p{2cm}<{\centering}|p{2cm}<{\centering}|p{2cm}<{\centering}}
			\toprule[1pt]
			($\lambda_1$)& 200&  20& 2000&200 &200  \\
			($\lambda_2$)& 0.002&  0.002& 0.002&0.02 &0.0002  \\
			\hline
			$L_2$ ($\downarrow$) &\textbf{28.32} &31.84&29.27&31.99 &29.10\\
			\hline
			Std ($\uparrow$) &\textbf{1.663}&0.720&1.577&0.749 &1.582\\
			\bottomrule[1pt]
	\end{tabular*}}
	\label{tab:para}
\end{table}

\subsection{Hyper-parameters Analysis}
\label{sec53}
There are two important hyper-parameters for the training of ACT-VAE, which are $\lambda_1$ and $\lambda_2$ in Eq.~\eqref{eqloss}. 
To analyze the influence of $\lambda_1$ and $\lambda_2$ in training, we conduct experiments with different values for $\lambda_1$ and $\lambda_2$. The value of the pair ($\lambda_1$, $\lambda_2$) in above experiments is (200, 0.002). 
In this section, we set its value as (20, 0.002), (2000, 0.002), (200, 0.02), (200, 0.0002) respectively. The results on Penn-action dataset can be seen in Table \ref{tab:para}, which shows that the value setting (200, 0.002) is rational. 
And users can adopt our value setting of $\lambda_1$ and $\lambda_2$, as an appropriate reference for different datasets.
The settings of $\lambda_3 \sim \lambda_6$ are based on hyper-parameters of pose-guided image generation methods (i.e., \cite{zhu2019progressive}). 

\subsection{Visual Analysis for Foreground Attention}
\label{sec8}
Since we adopt a foreground-background composition strategy for the P2I network, the P2I network has two types of outputs which are $\bar{I}_{t+i}$ and $M_{t+i}$ as shown in Eq.~\eqref{eqim}.
We show several cases about the $M_{t+i}$ in Fig.~\ref{fig:mask}, which illustrates that P2I network can distinguish the foreground and background in unsupervised learning, without the ground truth of mask.

\section{Conclusion}
We have proposed an effective framework for human action video prediction from a still image within various action categories. 
In our framework, ACT-VAE predicts pose by modeling the motion patterns and diversity in future videos with temporal coherence for each action category.
The temporal coherence is ensured by sampling the latent variable at each time step based on both historical latent variables and pose during inference.
When connected with a plug-and-play P2I network, ACT-VAE can synthesize image sequences and control action types in synthesis.
Extensive experiments on datasets containing complicated action videos illustrate the superiority of our framework.

\ifCLASSOPTIONcompsoc
\fi


\ifCLASSOPTIONcaptionsoff
\newpage
\fi



%


\bibliographystyle{IEEEtran}
\bibliography{egbib}

%

\end{document}